\documentclass[10pt,twocolumn,letterpaper]{article}

\usepackage[pagenumbers]{cvpr} 
\usepackage[accsupp]{axessibility}  

\definecolor{beierblue}{RGB}{0, 102, 204}
\definecolor{lightred}{RGB}{240,128,128}

\newcommand{\bx}{\mathbf{x}}
\newcommand{\bv}{\mathbf{v}}
\newcommand{\bt}{\mathbf{t}}
\newcommand{\by}{\mathbf{y}}

\newcommand{\ba}{\mathbf{a}}

\definecolor{cvprblue}{rgb}{0.21,0.49,0.74}
\usepackage[pagebackref,breaklinks,colorlinks,allcolors=cvprblue]{hyperref}
\usepackage{multirow}
\usepackage{colortbl}
\newcommand*\samethanks[1][\value{footnote}]{\footnotemark[#1]}
\def\paperID{***} 
\def\confName{CVPR}
\def\confYear{2025}

\title{Devils in Middle Layers of Large Vision-Language Models: Interpreting, Detecting and Mitigating Object Hallucinations via Attention Lens}

\author{Zhangqi Jiang\textsuperscript{1}
\quad
Junkai Chen\textsuperscript{2,3}
\quad
Beier Zhu\textsuperscript{4}
\quad
Tingjin Luo\textsuperscript{1}\thanks{Corresponding author.} 
\quad
Yankun Shen\textsuperscript{2,3}
\quad
Xu Yang\textsuperscript{2,3}\samethanks\\
\textsuperscript{1}National University of Defense Technology
\quad
\textsuperscript{2}Southeast University\\
\textsuperscript{3}Key Laboratory of New Generation Artificial Intelligence Technology \&\\
Its Interdisciplinary Applications (Southeast University), Ministry of Education\\
\textsuperscript{4}Nanyang Technological University\\
{\tt\small \{sxdxjzq,\ junkai.chen.0917\}@gmail.com
\quad
beier.zhu@ntu.edu.sg
\quad
tingjinluo@hotmail.com
}\\
{\tt\small \{220242353,\ xuyang\_palm\}@seu.edu.cn
}
}

\begin{document}
\maketitle
\begin{abstract}
Hallucinations in Large Vision-Language Models (LVLMs) significantly undermine their reliability, motivating researchers to explore the causes of hallucination. However, most studies primarily focus on the language aspect rather than the visual. In this paper, we address how LVLMs process visual information and whether this process causes hallucination. Firstly, we use the attention lens to identify the stages at which LVLMs handle visual data, discovering that the middle layers are crucial. Moreover, we find that these layers can be further divided into two stages: ``visual information enrichment'' and ``semantic refinement'' which respectively propagate visual data to object tokens and interpret it through text. By analyzing attention patterns during the visual information enrichment stage, we find that real tokens consistently receive higher attention weights than hallucinated ones, serving as a strong indicator of hallucination. Further examination of multi-head attention maps reveals that hallucination tokens often result from heads interacting with inconsistent objects. Based on these insights, we propose a simple inference-time method that adjusts visual attention by integrating information across various heads.
Extensive experiments demonstrate that this approach effectively mitigates hallucinations in mainstream LVLMs without additional training costs.\footnote{Code: \url{https://github.com/ZhangqiJiang07/middle_layers_indicating_hallucinations}.}
\end{abstract}    
\section{Introduction}
\label{sec:intro}

\begin{figure}[t]
    \centering
    \includegraphics[width=0.9\linewidth]{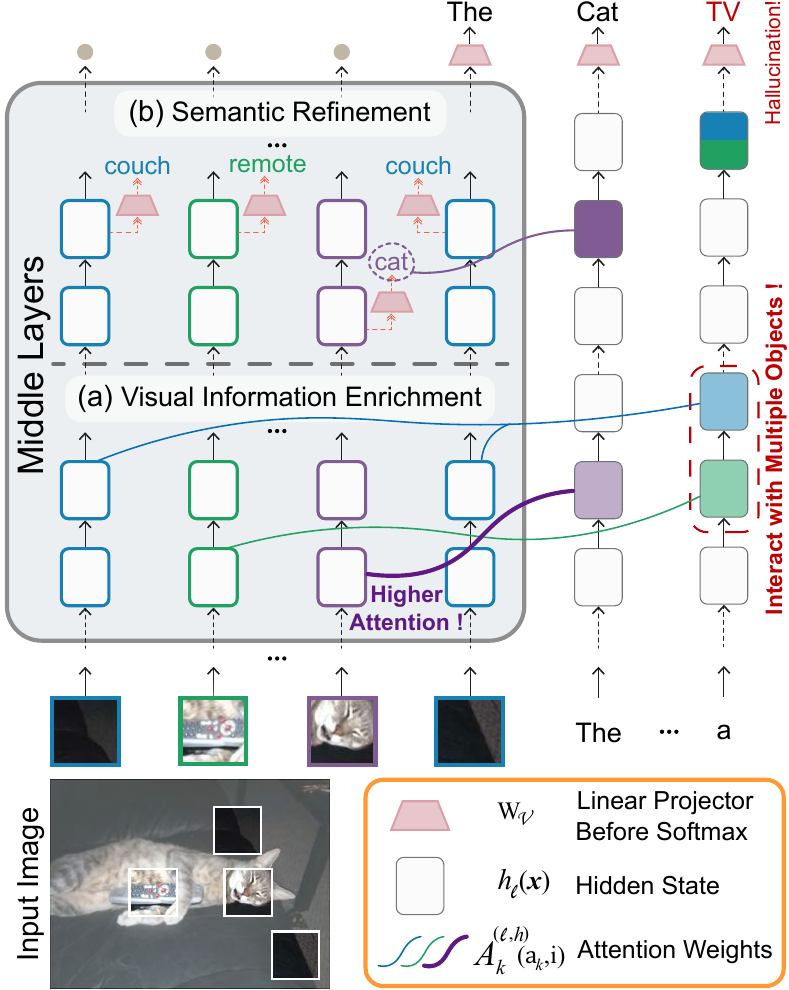}
    \caption{Illustration of our findings: (i) visual information is primarily processed in the middle layers where (a) the model extracts the visual information and then (b) interprets the semantics embedded in image tokens; (ii) for real tokens like ``cat'', the attention weights over image tokens are generally higher than hallucinated ones like ``TV'' in (a); (iii) the model may combine the visual features extracted from multiple objects to produce hallucinations.}
    \label{fig:illustrator}
\end{figure}

Building on the foundations of Large Language Models (LLMs), Large Vision-Language Models (LVLMs)~\cite{liu2024llava1.5,zhang2024wings,chen2023shikra,zhu2024minigpt,yin2023survey,peng2025lmm,zhao2024harmonizing} have emerged as powerful tools for understanding visual inputs and generating textual responses aligned with user intent. Their multi-modal nature has enabled a large variety of vision-language tasks, \eg, image captioning, visual question answering, and visual dialog. Despite generating coherent textual responses, these models often exhibit undesirable \textit{object hallucinations} -- producing objects that do not exist in the visual input.

Although various approaches, such as visual instruction fine-tuning~\cite{liu2024mitigating,jiang2024hallucination,gunjal2024detecting,yu2024hallucidoctor}, the integration of external expert models~\cite{yin2023woodpecker,chen2024halc,zhao2024mitigating,wu2024logical}, and contrastive decoding strategies~\cite{leng2024VCD,favero2024multi,wang2024mitigating,liu2024pai}, have been proposed to address the object hallucination issue in LVLMs, the underlying mechanisms driving these hallucinations remain poorly understood. A few studies have preliminarily explored the causes of object hallucination, identifying the \textbf{language bias} as the primary factor, with examples including the ``anchor pattern''~\cite{huang2024opera}  and ``text inertia''~\cite{liu2024pai}. In essence, these studies found that LVLMs tend to prioritize internal textual knowledge over external visual information as more tokens are generated.
However, these studies have overlooked the devils in the visual parts -- while intuitively, improper processing of visual information within the LVLMs could contribute more to the emergence of object hallucinations.

In this paper, we delve into how LVLMs process visual information from image tokens and how this affects the generation of object hallucinations. Building on the three-stage mechanism identified in LLMs for retrieving factual knowledge~\cite{geva2023dissecting}, we examine whether LVLMs exhibit similar distinct stages during inference and, if so, which stage is critical for processing visual information. 
We propose a simple score called Visual Attention Ratio (VAR) to examine the distribution of visual attention across layers, revealing that \textbf{the middle layers play a crucial role in processing visual information}.
To study how the model processes visual information in these middle layers, we leverage the logit lens~\cite{logitLens} method to decode the hidden states of image tokens into LVLMs' vocabulary, identifying two crucial stages as illustrated in~\cref{fig:illustrator}: a \textit{visual information enrichment} stage propagates visual information to the object token, and a \textit{semantic refinement} stage interprets the encoded visual information through text.

Secondly, focusing on these middle layers, we observe that the hallucinated tokens often exhibit (1) less active attention patterns, and (2) inconsistent attentions across different heads during \textit{visual information enrichment} stage, compared to real ones.
To quantify these observations, we first introduce a metric based on VAR, and confirm that \textbf{real object tokens tend to assign higher attention weights over image tokens during visual information enrichment than hallucinated ones}, illustrated in~\cref{fig:illustrator}.
Relying on our finding, we utilize the metric in this stage to detect hallucinated object tokens, achieving AUROC and mAP up to 74\% and 88\% on LLaVA-1.5-7B~\cite{liu2024llava1.5}.
Second, given the various attention heads used in the Multi-Heads Self-Attention (MHSA) mechanism~\cite{vaswani2017attention} for information aggregation, we explore the head behavior by visualizing the attention heatmaps over the input image. As illustrated in~\cref{fig:illustrator}, we find that \textbf{the heads interact with inconsistent objects during visual information enrichment when generating hallucinations}.

Our empirical findings provide valuable insights into the way object hallucinations are generated, intuitively described as follows (\cref{fig:illustrator}): In visual information enrichment, the model's ambiguous interactions with multiple objects result in limited and mixed visual information being propagated to the object token. Subsequently, in semantic refinement, this imprecise visual information leads the model to mistakenly associate semantic information from different objects, ultimately causing hallucinations.

Inspired by our findings, we propose a simple method to adjust the visual information process in the middle layers by integrating attention information from various heads during inference.
Extensive experiments on three mainstream LVLMs show significant hallucination mitigation compared to the original models, with average reductions in CHAIR$_I$ and CHAIR$_S$ up to 6.3 and 24.1 points, respectively, while preserving detail in descriptions.
\section{Related Work}
\label{sec:related_work}

\noindent\textbf{Large Vision-Language Models.}
The recent integration of advanced open-source LLMs, like Llama~\cite{touvron2023llama,touvron2023llama2} and Vicuna~\cite{chiang2023vicuna}, has greatly expanded the capabilities of LVLMs, enabling them to perform more complex vision-language tasks.
Typically, the architecture of recent LVLMs incorporates three key components: a vision branch, such as CLIP~\cite{radford2021clip} and EVA~\cite{fang2023eva}, to encode image inputs, a modality connector to transform image features into image embeddings aligning with text modality, and a pretrained language model to process image and prompt embeddings to generate responses.
LLaVA-1.5~\cite{liu2024llava1.5} employs an MLP to map the output of the vision branch into 576 image embeddings.
Similarly, Shikra~\cite{chen2023shikra} utilizes one linear layer to align image features.
While MiniGPT-4~\cite{zhu2024minigpt} adopts a learnable querying transformer to establish vision-language connections, with just 32 image tokens as LVLM input.
Despite their impressive capabilities, the above LVLMs suffer from severe object hallucinations. In this paper, we focus on interpreting, detecting, and mitigating their hallucinations.

\noindent\textbf{Object Hallucination in LVLMs.}
Object hallucination~\cite{zhou2024analyzing,zhai2024halleControl} is a prevalent and critical issue in current LVLMs where the model erroneously generates descriptions of non-existent objects in images, posing significant risks in high-stakes fields such as medical imaging~\cite{he2023geometric,kong2024multi}, sequential recommendation systems~\cite{guo2024scaling,shen2024optimizing}, and autonomous driving~\cite{cui2024survey}.
Previous studies to mitigate this issue have primarily focused on visual instruction fine-tuning~\cite{liu2024mitigating,jiang2024hallucination,gunjal2024detecting,yu2024hallucidoctor,zhu2023debiased}, incorporating external expert models~\cite{yin2023woodpecker,chen2024halc,zhao2024mitigating,wu2024logical}, and improving decoding strategies~\cite{huang2024opera,leng2024VCD,liu2024pai,wang2024mitigating,favero2024multi}.
Despite numerous efforts, our understanding of the underlying mechanism driving these errors remains limited.
Recent studies like OPERA~\cite{huang2024opera} have pointed specific ``anchor patterns'' in text tokens that always align with the onset of hallucinated contents, and PAI~\cite{liu2024pai} has identified ``text inertia'' as another contributing factor, as LVLMs reproduce identical hallucinated descriptions when solely input part of historical generated text.
Contrary to these perspectives that focus on language bias, we investigate the causes of object hallucinations from the view of visual information processing through the lens of attention.

\noindent\textbf{Interpretability in Foundation Models.}
In the field of NLP, numerous studies have analyzed LLMs to explore the internal model knowledge and understand model behavior in specific settings from the perspectives of attention maps~\cite{chefer2021generic,Yuksekgonul2024attention}, neural activation patterns~\cite{he2024llm}, and intermediate hidden states~\cite{su2024unsupervised,ghandehariounpatchscopes}.
The logit lens~\cite{logitLens}, which transforms the hidden states at each layer into the vocabulary space using the model's own linear projector before softmax, offers insightful observations into the next token prediction process~\cite{geva2022transformer,halawioverthinking} in LLMs.
Our work adapts the logit lens method to analyze hidden states of image tokens, enabling us to interpret the LVLMs' process of understanding the visual information via textual vocabulary.
Few studies explore the internal mechanisms of LVLMs~\cite{cao2020behind,cao2024empirical,gandelsman2024interpreting}, notable efforts include~\cite{palit2023towards}, which extends an unimodal causal tracing tool for studying neural mechanisms in BLIP's image-conditioned text generation, and~\cite{parekh2024concept}, which introduces a dictionary learning-based framework to extract multi-modal concepts for interpreting intermediate representations.
Compared to prior works, our study leverages the internal signals from attention weights to uncover the generation of hallucinated objects.
\section{Understanding Object Hallucinations}\label{sec:findings}
In this section, we start by introducing the generation process of LVLMs and the analytical tools, including visual attention ratio and logit lens. We then conduct case studies focusing on three key aspects: the processing of visual information, the attention patterns of object tokens, and the behavior of heads during object token generation.

\subsection{Preliminary}\label{subsec:preliminary}
\textbf{Notation.} The input to LVLMs is initialized with a sequence of  $d$-dimensional image tokens $\{\bv_1, ..., \bv_n\}$ and instruction text (prompt) tokens  $\{\bt_1, ...,\bt_m\}$. 
Typically structured as a Transformer decoder~\cite{vaswani2017attention}, the LVLM outputs responses in an autoregressive manner:
 at time step $k$, the model processes the initial input tokens $\{\bv_1, ..., \bv_n,\bt_1, ...,\bt_m\}$, followed by the sequence of $k{-}1$ previously generated tokens  $\{\by_1, ..., \by_{k-1}\}$, to predict the next token  $\by_k$. 
Let $L$ denote the number of Transformer layers, each with $H$ heads.
At layer $\ell {\in} [L]$, we denote the attention weights in head $h {\in} [H]$ as $\mathbf{A}_{k}^{(\ell,h)} {\in} \mathbb{R}^{a_k \times a_k}$, where $a_k{=}n{+}m{+}k{-}1$.
Additionally, we define $\bv^{\ell}_i$ and $\by^{\ell}_{k-1}$ as the hidden states for the image token $\bv_i$ and generated token $\by_{k-1}$ at layer $\ell$, respectively.
The real and hallucinated object tokens are noted as $\by_{\mathsf{r}}$ and $\by_{\mathsf{h}}$, respectively.

\noindent\textbf{Visual Attention Ratio.} For the $k$-th token $\by_k$, we define the Visual Attention Ratio (VAR) in each head $h$ at layer $\ell$:
\begin{equation}
    \textrm{VAR}^{(\ell,h)}(\by_k) \triangleq \sum_{i=1}^{n} \mathbf{A}_{k}^{(\ell,h)}(a_k, i), \label{eq:var}
\end{equation}
where $\mathbf{A}_{k}^{(\ell,h)}(a_k, i)$ represents the attention weights of the newly generated token $\by_k$ assigned to the image token $\bv_i$.
VAR quantifies the extent of the token $\by_k$’s interaction with visual information: \textit{a higher VAR score indicating a greater contribution from image tokens during $\by_k$’s generation}.

\noindent\textbf{Logit Lens.} We utilize this technique to investigate how the model interprets the visual hidden state $\bv_i^\ell$ during inference via text. 
We denote the linear projector before softmax as $\mathbf{W}_{\mathcal{V}} {\in} \mathbb{R}^{|\mathcal{V}| \times d}$, which is applied to $\by^L_{k-1}$ to predict the probability of next token over the vocabulary $\mathcal{V}$. Then, the logit lens transforms the hidden state $\bv_i^\ell$ of the image tokens to the prediction probability distribution over the vocabulary by the linear projector $\mathbf{W}_{\mathcal{V}}$:
\begin{align}
     \mathbf{p}(\mathcal{V}|\bv_i^\ell)  = \text{softmax}(\mathbf{W}_{\mathcal{V}} \cdot \bv_i^\ell) \in \mathbb{R}^{|\mathcal{V}|},
     \label{eq:logit_lens}
\end{align}
where $\mathbf{p}_j(\mathcal{V}|\bv_i^\ell)$ corresponds to the $j$-th text token in the vocabulary.
To explain the processed image token, the text token with the highest probability is considered as the model's interpretation of the hidden state $\bv_i^\ell$ at layer $\ell$.

\begin{figure}[t]
    \centering
    \includegraphics[width=1\linewidth]{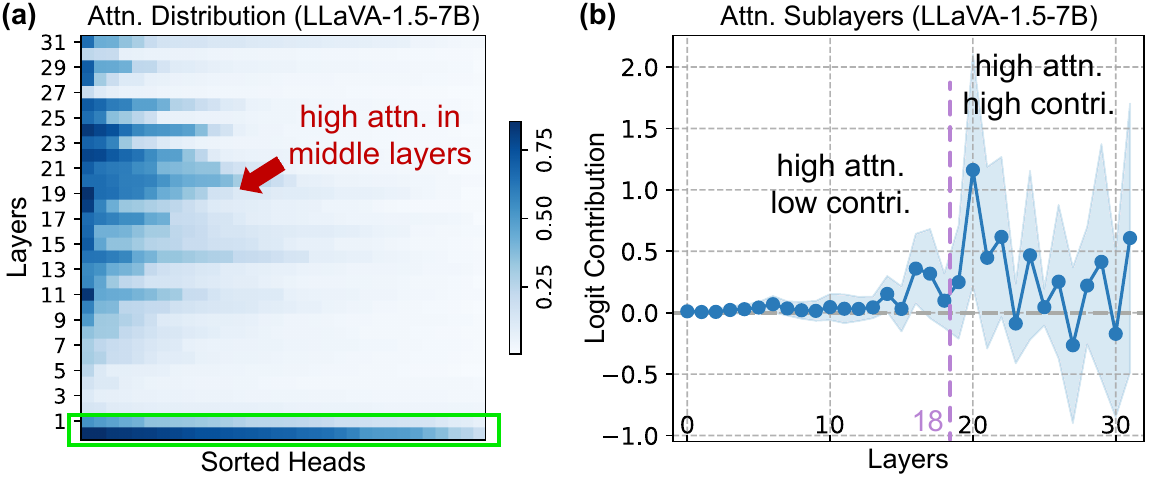} 
    \caption{(a) Distribution of visual attention ratio for real object tokens across heads and layers in LLaVA-1.5-7B, sorted row-wise by attention ratios. Note that the high attention in the $0$-th layer (bottom row, green rectangle) is not consistent across all LVLMs, as Shikra and MiniGPT-4 models fail to exhibit this pattern, see~\cref{appendix:case_studies}. (b) The logit contribution of attention sublayers to real token prediction. We find the middle layers continuously assign higher attention weights to image tokens and exhibit two different contribution patterns to the correct prediction.}
    \label{fig1:attention}
\end{figure}

\begin{figure*}[thb]
    \centering
    \includegraphics[width=1\linewidth]{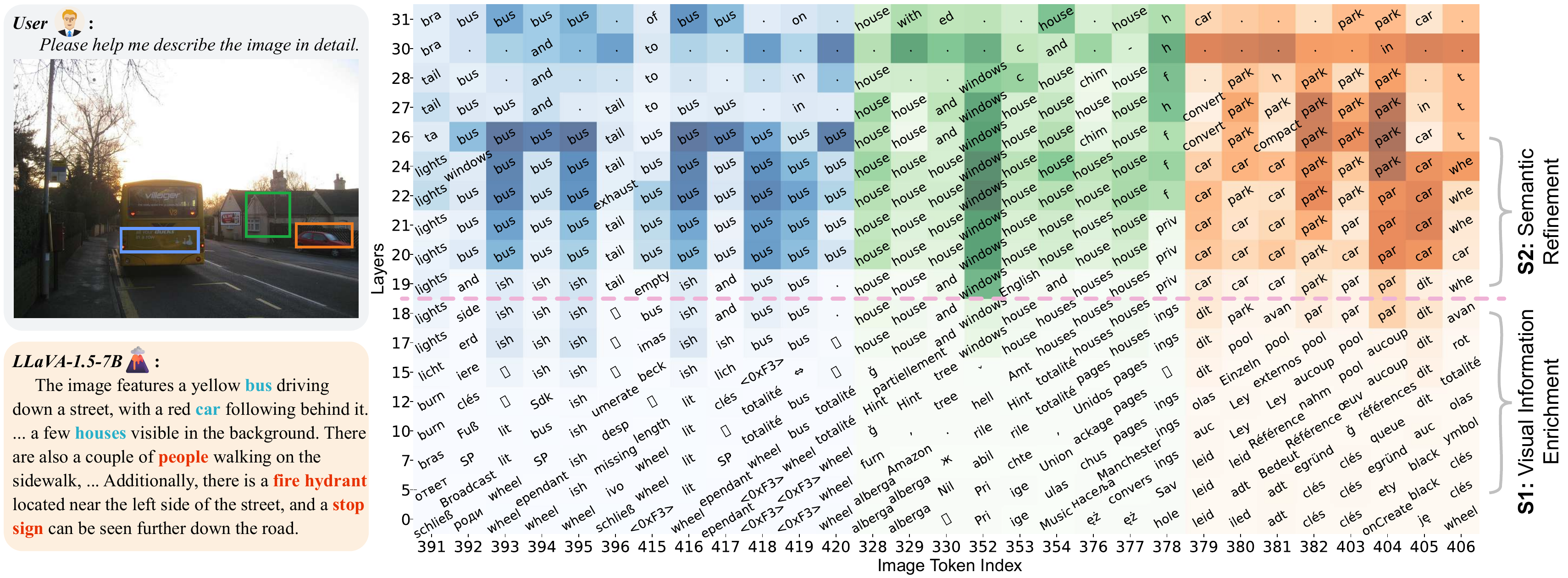}
    \caption{Case study of image hidden state interpretation in LLaVA-1.5-7B via logit lens. The heatmap illustrates the retrieved texts of the image hidden states across layers in the three distinct regions, differentiated by color. Our findings reveal that the model's semantic comprehension of image tokens emerges in the later middle layers (19-26) while remaining largely absent in earlier layers. The real and hallucinated object tokens are presented in blue and red in the description, respectively.}
    \label{fig2:logit_lens}
\end{figure*}

\subsection{Experimental Setup for Case Study} %
We conduct case studies with a randomly selected subset of 2,000 images from COCO 2014 validation set~\cite{lin2014mscoco}.
These images represent diverse scenes and activities across 80 common object categories, each paired with multiple object annotations.
The 7B version of LLaVA-1.5 serves as the main focus for the subsequent analysis. In~\cref{appendix:case_studies}, we further repeat our case studies on other LVLMs, including Shikra and MiniGPT-4, and observe consistent findings.

To investigate the internal patterns in generating real and hallucinated object tokens, we use the model with greedy search to generate captions for the selected images, prompted by ``\textit{Please help me describe the image in detail}.''. Real and hallucinated object tokens are then identified using ground truth annotations as a reference. For multi-token objects, only the first token is considered. As a result, we obtained 1,842 hallucinated tokens and 4,397 real tokens.

\subsection{Finding 1: Middle Layers Matter for Visual Information Interaction}\label{finding1}
To assess the contribution of visual information to token generations, we compute the VAR score (\cref{eq:var}) for real object tokens $\by_{\mathsf{r}}$ across all layers $\ell {\in} [L]$ and heads $h {\in} [H]$. We calculate the mean of $\textrm{VAR}^{(\ell,h)}(\by_{\mathsf{r}})$ for all real object tokens in our case dataset, with the results depicted in~\cref{fig1:attention}~(a). The visualization shows that the middle layers, \ie, 5-26 layers, continuously exhibit high attention weights to the image tokens, indicating that \textbf{visual information interaction primarily occurs in the middle layers}.

We further analyze how the model processes visual information by examining signals from the MHSA sublayers. Using the logit lens, we quantify the contribution of the MHSA sublayers to real token prediction as $\mathbf{p}_\mathsf{o}(\mathcal{V}|\ba^\ell_{\mathsf{r}})$, where $\ba^\ell_{\mathsf{r}}$ is the output of the MHSA sublayer at layer $\ell$, and $\mathsf{o}$ denotes the index of the real object token $\by_{\mathsf{r}}$ in the vocabulary.
\cref{fig1:attention}~(b) reveals two patterns in the contributions from the MHSA sublayers: lower in 5-18 layers and higher in 19-26 layers, reflecting varied use of visual information.

To better understand these two patterns, we use the logit lens to map the hidden states of image token $\bv_i^\ell$ to text. Specifically, for each image token $\bv_i$ at layer $\ell$, we retrieve the text token in the vocabulary with the highest probability, \ie, $\textrm{argmax}_{1 \leq j \leq |\mathcal{V}|} \{\mathbf{p}_j(\mathcal{V}|\bv_i^\ell)\}$.
A case is shown in~\cref{fig2:logit_lens}, revealing two stages in the middle layers with different visual information usage:
\begin{itemize}
    \item \textbf{Stage 1: Visual Information Enrichment}. The first stage, in layers 5–18, shows that retrieved text tokens are less related to the corresponding image patches, suggesting the model is unable to interpret visual information. This misalignment might explain the low prediction contributions observed in~\cref{fig1:attention}~(b). Meanwhile, these layers continue to show high VAR scores, indicating that the object tokens are accumulating visual information through self-attention. In~\cref{finding2,finding3}, we show that such attention serves as a key indicator to identify hallucinations. 
   \item \textbf{Stage 2: Semantic Refinement}. The second stage, in layers 19–26, shows that the retrieved text tokens are semantically consistent with the image patches, suggesting the model is able to interpret and utilize the visual information encoded in the image tokens. Given the high VAR during this stage, we term it as semantic refinement where the model actively interacts with the semantic information of image tokens to reason object token prediction.
\end{itemize}

\subsection{Finding 2: Inactive Visual Attention in Middle Layers Implies Hallucination}\label{finding2}
\begin{figure}[t]
    \centering
    \includegraphics[width=1\linewidth]{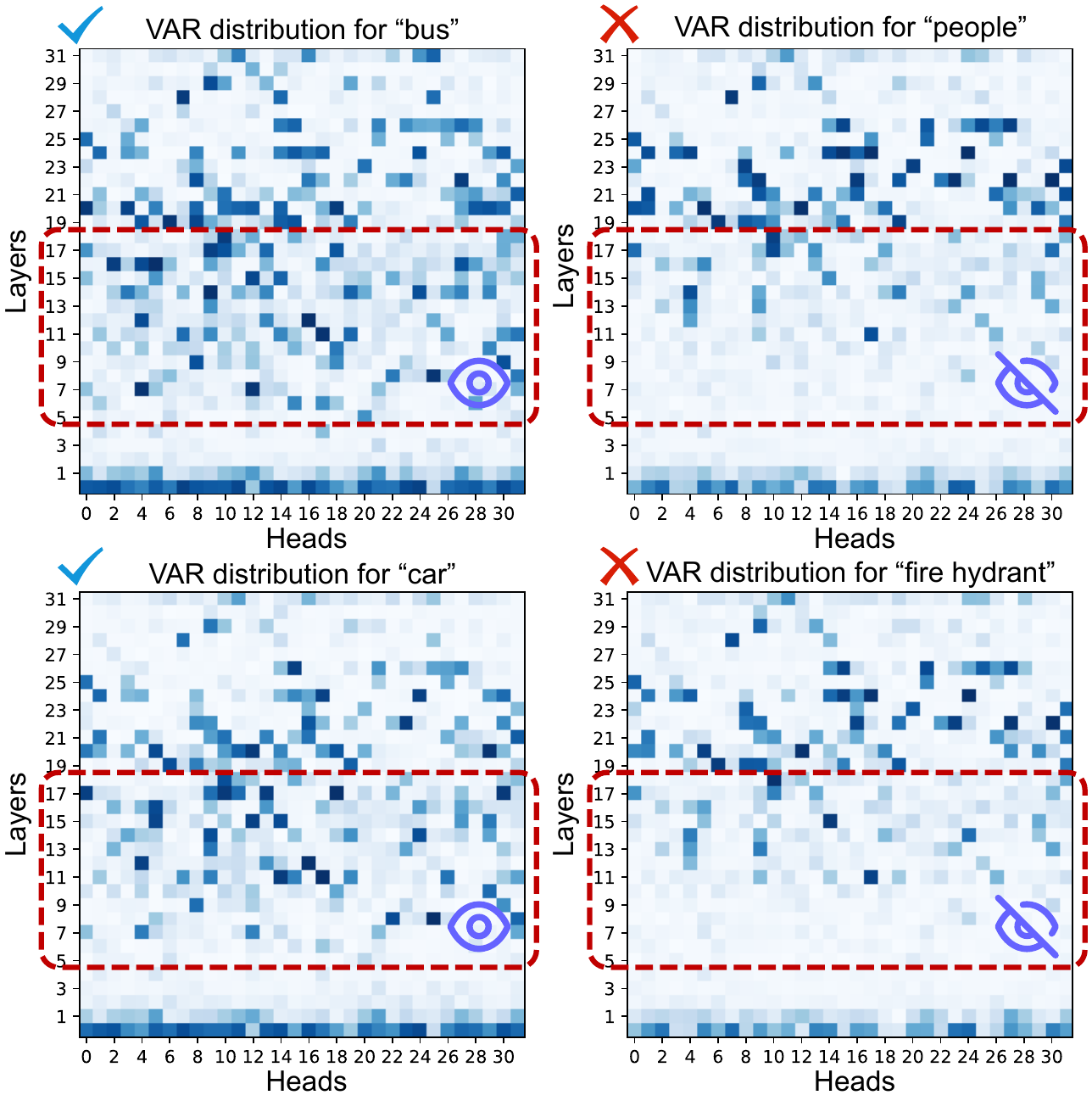}
    \caption{Case study of the visual attention ratio distribution over heads and layers in LLaVA-1.5-7B. Left: real objects; Right: hallucinated objects. We find that the hallucinated object tokens exhibit inactive attention patterns during visual information enrichment compared to the real ones.}
    \label{fig3:O2_case}
\end{figure}
Given that visual information is primarily processed in the middle layers, a natural question arises: does the model display unusual attention patterns in these layers when generating hallucinated object tokens?
\cref{fig3:O2_case} presents the VAR distribution for real object tokens (`bus' and `car') and hallucinated ones (`people' and `fire hydrant') from the case of~\cref{fig2:logit_lens}. We observe that the hallucinated object tokens demonstrate an inactive attention pattern during the visual information enrichment stage (5-18 layers) compared to the real ones.
We conjecture that such pattern weakens the interaction between the object token and image tokens in this stage, limiting the propagation of visual information and potentially leading to hallucinations.
To quantify this observation, we introduce a metric, Summed Visual Attention Ratio (SVAR), calculated by averaging VAR scores over all heads and summing across selected layers $[\ell_s, \ell_e]$. Specifically, for a real object token $\by_{\mathsf{r}}$ in $[\ell_5,\ell_{18}]$, we compute:
\begin{equation}
    \textrm{SVAR}_{5\textrm{-}18}(\by_{\mathsf{r}}) \triangleq \frac{1}{H} \sum_{\ell = 5}^{18} \sum_{h=1}^{H} \textrm{VAR}^{(\ell, h)}(\by_{\mathsf{r}}). \label{eq:svar_5_18}
\end{equation}
We conduct a statistical experiment on the $\textrm{SVAR}_{5\textrm{-}18}$ metric for both real and hallucinated tokens.
The results, depicted in~\cref{fig:statistic}~(a), and the associated statistical test detailed in~\cref{appendix:statistical_test}, indicate a salient difference: \textbf{the model assigns significantly higher attention weights to image tokens when generating real object tokens during visual information enrichment compared to hallucinated ones.}
\cref{fig:statistic}~(b) shows a similar trend in LLaVA-1.5-13B.
\begin{figure}[t]
    \centering
    \includegraphics[width=.95\linewidth]{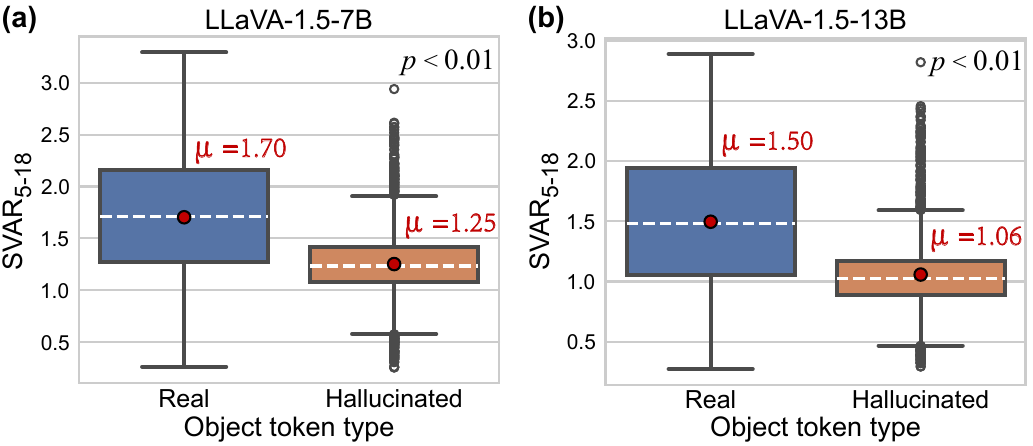}
    \caption{$\textrm{SVAR}_{5\textrm{-}18}$ score distributions across object token types for the 7B (a) and 13B (b) versions of LLaVA-1.5.}
    \label{fig:statistic}
\end{figure}

\subsubsection{Application: Object Hallucination Detection} \label{subsec:detection}
Building upon our Finding 2, we apply the $\textrm{SVAR}_{5\textrm{-}18}$ metric to detect object hallucinations.
We assess the efficacy of $\textrm{SVAR}_{5\textrm{-}18}$ in reflecting object hallucination using our case dataset, where a positive sample is a real object token and a negative sample is a hallucinated one.
For comparison, we use the internal confidence metric recently proposed in~\cite{jiang2024interpreting} as a baseline.
The internal confidence uses the maximum probability of the object token within all image hidden states $\{\bv_i^\ell: i {\in} [n], \ell {\in} [L]\}$, mapped by the logit lens for detection.
The qualitative results are presented in~\cref{fig5:roc_pr_curve} and the results for LLaVA-1.5-13B are shown in~\cref{appendix:case_studies}.
Compared to internal confidence, using the simple $\textrm{SVAR}_{5\textrm{-}18}$ metric improves the AUROC by 8.82\% and the AP by 3.53\%, verifying the utility of our metric.
This further suggests that the middle layers provide crucial indicator information for object hallucination.

\begin{figure}[t]
    \centering
    \includegraphics[width=.95\linewidth]{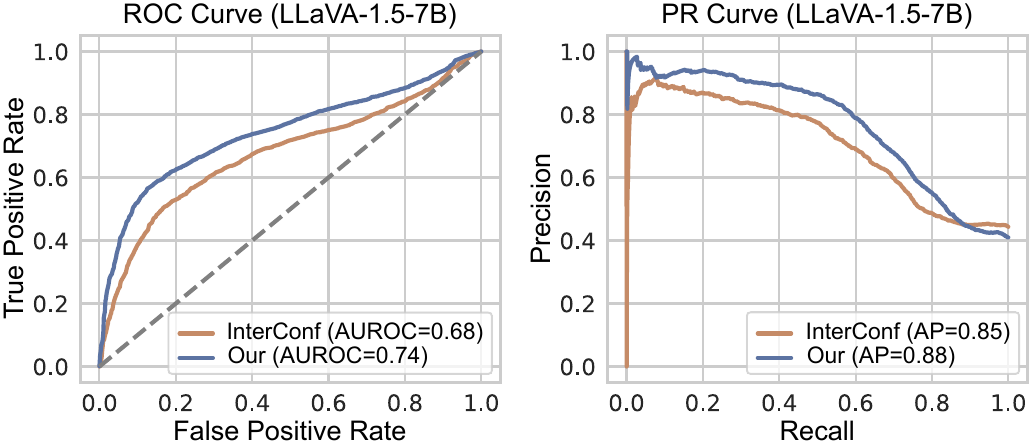}
    \caption{Object hallucination detection curves for LLaVA-1.5-7B. We show the ROC and Precision-Recall curves of $\textrm{SVAR}_{5\textrm{-}18}$ metric for object hallucination detection on the case study dataset.}
    \label{fig5:roc_pr_curve}
\end{figure}

\begin{figure*}[t]
    \centering
    \includegraphics[width=.97\linewidth]{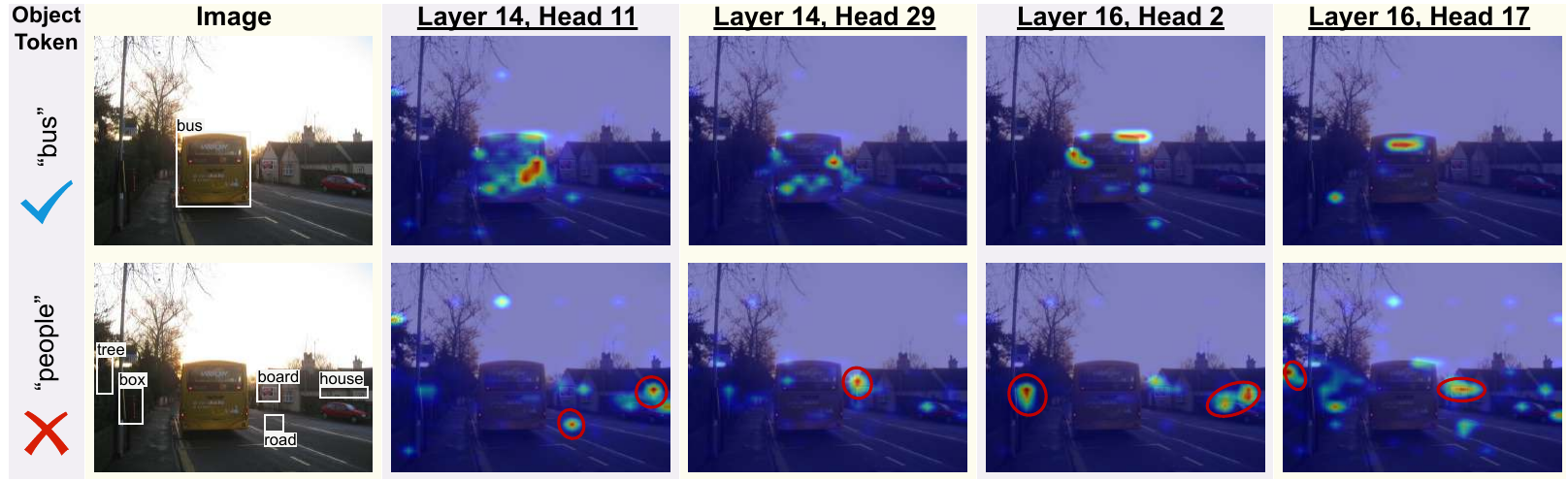}
    \caption{Visualization of attention maps over image for `bus' (real) and `people' (hallucinated) object tokens in heads $(\ell, h)$. We find that the heads tend to interact with multiple objects when generating hallucinated object tokens. More examples can be found in~\cref{appendix:head_behavior}.}
    \label{fig:heads_heatmap}
\end{figure*}

Beyond the hand-crafted $\textrm{SVAR}_{5\textrm{-}18}$ measure, we explore the potential of middle layers to indicate hallucinations by training a simple model to derive metrics directly from raw visual attention states. 

\noindent\textbf{Experimental setting.} 
We train a two-layer MLP by inputting the concatenated VAR scores across all heads in the layer range of interest $[\ell_s,\ell_e]$ to classify the real and hallucinated object tokens. Specifically, the input vector $\bx(\by_{\mathsf{r}})$ for the real token $\by_{\mathsf{r}}$ can be written as:
\begin{equation}
    \bx(\by_{\mathsf{r}})=[\textrm{VAR}^{(\ell_s,1)}, \textrm{VAR}^{(\ell_s,2)}, \dots, \textrm{VAR}^{(\ell_e,H)}]^\top.
\end{equation}
We split the case dataset into training and test sets in the ratio of 8:2, and train the MLP with cross-entropy loss. The model performance is evaluated using accuracy, AUROC, and recall metrics. Training details are in~\cref{appendix:detector_train}.

\noindent\textbf{Results.} We experiment with four layer ranges in LLaVA-1.5-7B: $[0,4]$, $[5,18]$, $[19,26]$, and $[27,31]$, as shown in~\cref{tab:layer_res}. We can see using the layers of visual information enrichment (5-18) achieves the best performance among other ranges, validating the effectiveness of our Finding 2. We observe that other layers also encode hallucination-related information which can be learned to detect hallucinations. However, compared to the \textit{explicit} pattern in 5-18 layers, the \textit{implicit} patterns contained in other ranges require computation costs for feature learning.

\subsection{Finding 3: Attention Heads Interact with Multiple Objects Incurring Object Hallucination}\label{finding3}
Considering the MHSA mechanism incorporates various attention heads to aggregate information, we analyze heatmaps over the image for both real and hallucinated object tokens (`bus' vs `people') to investigate the head behavior during visual information enrichment.
A case is shown in~\cref{fig:heads_heatmap} and more examples can be found in~\cref{appendix:head_behavior}.
We can see that each head focuses on distinct local details, yet the attention distribution for the real token consistently aligns with the spatial extent of the corresponding object.
\textbf{In contrast, for the hallucinated token, the heads interact with inconsistent objects in the image during this stage.}
We speculate that such unstable head behavior encodes mixed information from multiple objects into the object token, potentially leading to hallucinations.

\begin{table}[t]
    \centering
    \resizebox{.9\linewidth}{!}{
    \begin{tabular}{lcccc|c}
        \toprule
        \textbf{Layers} & \textbf{0-4} & \textbf{5-18} & \textbf{19-26} & \textbf{27-31} & \textbf{0-31} \\
        \midrule
        Accuracy & 82.61 & \textbf{84.46} & 81.49 & 77.72 & 84.46 \\
        AUROC & 89.57 & \textbf{90.19} & 87.34 & 84.30 & 90.28 \\
        Recall & 66.22 & \textbf{72.34} & 68.88 & 63.56 & 71.28 \\
        \bottomrule
    \end{tabular}
    }
    \caption{Object hallucination detection results (\%) of different layer ranges of interest on LLaVA-1.5-7B.}
    \label{tab:layer_res}
\end{table}

\section{Object Hallucination Mitigation}

\begin{table*}[t]
    \centering
    \resizebox{.92\linewidth}{!}{
    \begin{tabular}{lcccccccccccc|cc}
        \toprule
        \multirow{2}{*}{\textbf{Method}} &  \multicolumn{3}{c}{\textbf{LLaVA-1.5-7B}} & \multicolumn{3}{c}{\textbf{LLaVA-1.5-13B}} & \multicolumn{3}{c}{\textbf{Shikra-7B}} & \multicolumn{3}{c}{\textbf{MiniGPT-4-7B}} & \multicolumn{2}{c}{\textbf{Avg.}} \\
        \cmidrule{2-4}\cmidrule{5-7}\cmidrule{8-10}\cmidrule{11-13}\cmidrule{14-15}
         & $C_S\downarrow$ & $C_I\downarrow$ & F1$\uparrow$ & $C_S\downarrow$ & $C_I\downarrow$ & F1$\uparrow$ & $C_S\downarrow$ & $C_I\downarrow$ & F1$\uparrow$ & $C_S\downarrow$ & $C_I\downarrow$ & F1$\uparrow$ & $C_S\downarrow$ & $C_I\downarrow$ \\
        \midrule
        \multicolumn{15}{l}{\textit{Decoding Strategy}} \\
        \midrule
        Greedy & 53.0 & 15.6 & \cellcolor{gray!15}76.7 & 49.8 & 14.6 & \cellcolor{gray!15}78.2 & 57.6 & 15.7 & \cellcolor{gray!15}75.3 & 31.8 & 12.0 & \cellcolor{gray!15}71.1 & 48.1 & 14.5 \\
        Beam & 55.6 & 15.4 & \cellcolor{gray!15}77.5 & 50.4 & 13.8 & \cellcolor{gray!15}79.0 & 59.0 & 16.3 & \cellcolor{gray!15}74.6 & 29.2 & 9.9 & \cellcolor{gray!15}71.2 & 48.6 & 13.9 \\
        OPERA & 45.6 & 13.1 & \cellcolor{gray!15}79.1 & 42.6 & 13.2 & \cellcolor{gray!15}77.8 & 41.4 & 13.7 & \cellcolor{gray!15}73.5 & 25.4 & 9.6 & \cellcolor{gray!15}71.2 & 38.8 & 12.4 \\
        \midrule
        \multicolumn{15}{l}{\textit{Contrastive Decoding}} \\
        \midrule
        VCD$^{\dagger}$ & 58.6 & 18.2 & \cellcolor{gray!15}72.8 & 53.6 & 15.3 & \cellcolor{gray!15}75.8 & 56.4 & 15.5 & \cellcolor{gray!15}75.2 & 41.4 & 14.1 & \cellcolor{gray!15}68.2 & 52.5 & 15.8 \\
        PAI$^{\dagger}$ & \textbf{24.2} & 7.1 & \cellcolor{gray!15}75.2 & 33.0 & 9.2 & \cellcolor{gray!15}77.8 & 38.6 & 10.1 & \cellcolor{gray!15}76.2 & 23.2 & 8.2 & \cellcolor{gray!15}71.4 & 29.8 & 8.7 \\
        \midrule
        \textbf{Ours}$^{\dagger}$ & 25.0 & \textbf{6.7} & \cellcolor{gray!15}76.1 & \textbf{25.8} & \textbf{8.8} & \cellcolor{gray!15}77.3 & \textbf{23.8} & \textbf{9.4} & \cellcolor{gray!15}72.7 & \textbf{21.4} & \textbf{8.0} & \cellcolor{gray!15}70.8 & \textbf{24.0} & \textbf{8.2} \\
        $\Delta$\% & \textcolor{lightred}{$\uparrow$3.3\%} & \textcolor{cvprblue}{$\downarrow$5.6\%} &  & \textcolor{cvprblue}{$\downarrow$21.8\%} & \textcolor{cvprblue}{$\downarrow$4.3\%} &  & \textcolor{cvprblue}{$\downarrow$38.3\%} & \textcolor{cvprblue}{$\downarrow$6.9\%} &  & \textcolor{cvprblue}{$\downarrow$7.8\%} & \textcolor{cvprblue}{$\downarrow$2.4\%} &  & \textcolor{cvprblue}{$\downarrow$19.5\%} & \textcolor{cvprblue}{$\downarrow$5.7\%} \\
        \bottomrule
    \end{tabular}
    }
    \caption{CHAIR hallucination evaluation and F1 results on three LVLMs with \textit{max new token} set to 512. $\dagger$ denotes using the greedy decoding strategy. Our method outperforms other baselines and also preserves the richness of the descriptions. $\Delta$\% denotes the relative performance improvement with respect to the second-best method.}
    \label{tab:main_results}
\end{table*}

\subsection{Heads Guided Attention Intervention}\label{subsec:method}
Inspired by the insights from~\cref{sec:findings} that improper processing of visual information during inference can lead to object hallucinations, we aim to correct this process to alleviate object hallucination.
Specifically, we adjust the attention weights assigned to image tokens, \ie, $\{\mathbf{A}_{k}^{(\ell, h)}(a_k, i)\}_{i=1}^{n}$, in each head at the middle layers during inference for intervention.
To modify the attention weights in time step $k$, we extract the attention score matrix $\mathbf{S}_{k}^{(\ell, h)}$ before softmax operation:
\begin{equation}
    \mathbf{S}_{k}^{(\ell, h)} = \left(\frac{\mathbf{Q}_{\ell, h} \mathbf{K}_{\ell, h}^\top}{\sqrt{d_k}}\right)_k,
\end{equation}
where $\mathbf{Q}_{\ell, h}$ and $\mathbf{K}_{\ell, h} {\in} \mathbb{R}^{a_k \times d_k}$ represent the query and key matrices of dimension $d_k$, respectively.

Guided by the attention pattern difference in Finding 2, we amplify the original attention scores in the layers of visual information enrichment by adding positive values to enhance visual information interaction.
Furthermore, leveraging the head behavior observed in Finding 3, we compute these positive values by averaging the absolute attention scores across all heads, where only the consistent regions interacted by different heads receive large enhancement.
By integrating the attention information from various heads, we can find a more faithful and object-related direction for attention shift to reduce hallucinations.
Formally, for each image token $i {\in} [n]$ across all heads $h {\in} [H]$ in layer $\ell {\in} [\ell_s, \ell_e]$, we adjust the visual attention scores by:
\begin{equation}
    \mathbf{S}_{k}^{(\ell, h)}(a_k, i) = \mathbf{S}_{k}^{(\ell, h)}(a_k, i) + \alpha\frac{1}{H}\sum_{h=1}^H |\mathbf{S}_{k}^{(\ell, h)}(a_k, i)|, \label{eq:enhance}
\end{equation}
where the $[\ell_s, \ell_e]$ denotes the range of visual information enrichment, and the parameter $\alpha$ is defined as a balance factor to control the intervention strength.

\subsection{Experimental Setup and Results}
\noindent\textbf{Models.} 
We conduct experiments on three representative LVLMs to evaluate the effectiveness and generalization of our method, including the LLaVA-1.5~\cite{liu2024llava1.5}, Shikra~\cite{chen2023shikra}, and MiniGPT-4~\cite{zhu2024minigpt}.
In our experiments, we use the 7B and 13B versions of LLaVA-1.5 to study the scale effect, while other LVLMs are 7B models.

\noindent\textbf{Baselines.} We adopt two commonly used decoding strategies and three mainstream hallucination mitigation approaches as the baseline methods, including greedy decoding, beam search decoding~\cite{Ilya2014sequence}, OPERA~\cite{huang2024opera}, VCD~\cite{leng2024VCD}, and PAI~\cite{liu2024pai}. Greedy decoding selects the token with the highest probability over the vocabulary as the predicted next token. Beam search decoding maintains multiple beams and selects the top tokens with the highest cumulative probabilities at the end of generation. Improved on beam search, OPERA introduces an overtrust penalty term on the attention weights during inference with a rollback strategy to mitigate hallucinations. Different from the above three decoding approaches, VCD and PAI can be categorized as contrastive decoding methods. Specifically, VCD subtracts the output logits of the distorted visual input from the original output logits to reduce the statistical priors. In addition to using contrastive decoding, PAI further manipulates the attention matrix to overcome language bias. In our experiments, we use the default parameters of these baselines and unify $N_{\textrm{beam}}{=}5$ for the beam search decoding and OPERA.

\noindent\textbf{Benchmark and Metrics.} Following~\cite{huang2024opera}, we conduct experiments on 500 random images from the COCO 2014 validation set. To evaluate the degree of object hallucinations in the image captioning task, we adopt the CHAIR~\cite{rohrbach2018chair} criteria which computes the proportion of all objects mentioned in the caption that are not present in the ground-truth annotations. CHAIR provides two main metrics, including CHAIR$_I$ ($C_I$) and CHAIR$_S$ ($C_S$) that assess instance-level and sentence-level hallucinations, respectively:
\begin{equation}
    C_I = \frac{|\textrm{\scriptsize \{hallucinated objects\}}|}{|\textrm{\scriptsize \{all mentioned objects\}}|}, C_S = \frac{|\textrm{\scriptsize \{captions w/ hallucinated objects\}}|}{|\textrm{\scriptsize \{all captions\}}|}, \nonumber
\end{equation}
where lower values indicate fewer hallucinations. Following~\cite{liu2024pai}, we also adopt the F1 score to assess the richness and accuracy of the generated descriptions.

\noindent\textbf{Implementation details.} Our method contains two parameters: the visual information enrichment range $[\ell_s, \ell_e]$ and the balance factor $\alpha$ in~\cref{eq:enhance}.
Utilizing the VAR feature and logit lens described in~\cref{subsec:preliminary}, we can easily establish suitable ranges for any other LVLMs.
In our experiment, we set the ranges as $[5, 18]$ for both versions of LLaVA-1.5, $[3, 13]$ for Shikra, and $[3, 14]$ for MiniGPT-4. We set $\alpha$ to 0.5 for LLaVA-1.5, MiniGPT-4, and 0.55 for Shikra. We implement our method with greedy decoding as a baseline.

\noindent\textbf{Results.}
\cref{tab:main_results} presents the experimental results for the selected LVLMs with \textit{max new tokens} of 512.
From the \cref{tab:main_results}, our approach outperforms the three decoding strategies in reducing hallucinations and achieves an average reduction of 19.5\% in $C_S$ and 5.7\% in $C_I$, compared to the second-best baseline.
Compared to VCD, our method not only achieves superior hallucination mitigation performance but also better preserves the richness of the descriptions.
Compared to PAI which requires the cost of once additional forward process for contrastive decoding, our approach achieves comparable or superior performance by only adjusting attention weights during inference, validating the efficacy of our method.
Moreover, the consistent reduction of hallucinations across four different LVLMs confirms the generalizability of our findings.
To evaluate the impact of generated token lengths, we conduct experiments on LLaVA-1.5-7B by varying the \textit{max new token} in $\{64, 128, 256\}$. The results, reported in~\cref{tab:vary_max_new_token}, show that our method consistently reduces hallucinations across different token lengths and achieves the best performance on average, demonstrating its robustness. We provide qualitative results in~\cref{appendix:qualitative_results}.

\begin{table}[t]
    \centering
    \resizebox{\linewidth}{!}{
    \large
    \begin{tabular}{lccccccccc|cc}
        \toprule
        \multirow{2}{*}{\textbf{Method}} &  \multicolumn{3}{c}{\textbf{\textit{max new token} 64}} & \multicolumn{3}{c}{\textbf{\textit{max new token} 128}} & \multicolumn{3}{c}{\textbf{\textit{max new token} 256}}  & \multicolumn{2}{c}{\textbf{Avg.}} \\
        \cmidrule{2-4}\cmidrule{5-7}\cmidrule{8-10}\cmidrule{11-12}
         & $C_S\downarrow$ & $C_I\downarrow$ & F1$\uparrow$ & $C_S\downarrow$ & $C_I\downarrow$ & F1$\uparrow$ & $C_S\downarrow$ & $C_I\downarrow$ & F1$\uparrow$ & $C_S\downarrow$ & $C_I\downarrow$ \\
        \midrule
        \multicolumn{10}{l}{\textit{Decoding Strategy}} \\
        \midrule
        Greedy & 23.8 & 8.0 & \cellcolor{gray!15}74.7 & 52.4 & 15.5 & \cellcolor{gray!15}76.7 & 53.0 & 15.6 & \cellcolor{gray!15}76.7 & 43.1 & 13.0 \\
        Beam & \textbf{17.6} & \underline{6.0} & \cellcolor{gray!15}75.1 & 54.4 & 14.8 & \cellcolor{gray!15}77.6 & 55.6 & 15.4 & \cellcolor{gray!15}77.5 & 42.5 & 12.1 \\
        OPERA & 19.0 & 6.3 & \cellcolor{gray!15}74.4 & 44.2 & 12.9 & \cellcolor{gray!15}78.8 & 45.6 & 13.1 & \cellcolor{gray!15}79.1 & 36.3 & 10.8 \\
        \midrule
        \multicolumn{12}{l}{\textit{Contrastive Decoding}} \\
        \midrule
        VCD$^{\dagger}$ & 26.0 & 9.3 & \cellcolor{gray!15}73.3 & 56.8 & 16.9 & \cellcolor{gray!15}74.5 & 58.6 & 18.2 & \cellcolor{gray!15}72.8 & 47.1 & 14.8 \\
        PAI$^{\dagger}$ & 19.8 & 6.2 & \cellcolor{gray!15}74.0 & \textbf{24.2} & \underline{7.3} & \cellcolor{gray!15}75.2 & \textbf{24.2} & \underline{7.2} & \cellcolor{gray!15}75.2 & \textbf{22.7} & 6.9 \\
        \midrule
        \textbf{Ours}$^{\dagger}$ & \underline{18.2} & \textbf{5.4} & \cellcolor{gray!15}73.9 & \underline{24.8} & \textbf{7.1} & \cellcolor{gray!15}76.0 & \underline{25.0} & \textbf{7.1} & \cellcolor{gray!15}76.1 & \textbf{22.7} & \textbf{6.5} \\
        \bottomrule
    \end{tabular}
    }
    \caption{CHAIR hallucination evaluation and F1 results on LLaVA-1.5-7B with varying \textit{max new token} in $\{64, 128, 256\}$. $\dagger$ denotes using the greedy decoding strategy.}
    \label{tab:vary_max_new_token}
\end{table}
\section{Ablations and Discussions}\label{ablation}

\begin{table}[t]
    \centering
    \resizebox{.83\linewidth}{!}{
    \begin{tabular}{ccccc|c}
        \toprule
        \textbf{Layers} & \textbf{0-4} & \textbf{5-18} & \textbf{19-26} & \textbf{27-31} & \textbf{Greedy} \\
        \midrule
        $C_S\downarrow$ & 31.0 & \textbf{25.0} & 53.4 & 53.4 & 53.0 \\
        $C_I\downarrow$ & 10.4 & \textbf{6.7} & 14.3 & 15.7 & 15.6 \\
        F1$\uparrow$ & \cellcolor{gray!15}77.1 & \cellcolor{gray!15}76.1 & \cellcolor{gray!15}77.5 & \cellcolor{gray!15}77.2 & \cellcolor{gray!15}76.7 \\
        \bottomrule
    \end{tabular}
    }
    \caption{CHAIR hallucination evaluation and F1 score among four layer ranges on LLaVA-1.5-7B with \textit{max new token} set to 512.}
    \label{tab:ablation_middle_layer}
\end{table}

\noindent\textbf{Layers of attention intervention.}
Motivated by the empirical observation of implicit hallucination patterns across layers beyond the 5-18 range in~\cref{subsec:detection}, we investigate whether interventions in these layers can also mitigate object hallucinations.
We apply our approach (\cref{eq:enhance}) to the layer ranges $[0,4]$, $[19,26]$, and $[27,31]$ in LLaVA-1.5-7B, and the results are reported in~\cref{tab:ablation_middle_layer}.
We observe that the 5-18 layer range largely outperforms the other ranges while using some ranges (19-26 and 27-31) even enhanced hallucinations instead. The result suggests that explicit patterns in visual information enrichment are more effective than implicit patterns, facilitating more effective hallucination mitigation through simple attention intervention.

\noindent\textbf{Different attention intervention strategies.}
To validate the efficacy of our method in adjusting attention weights to reduce object hallucinations, we compare it with the inference intervention component proposed in PAI~\cite{liu2024pai}.
The inference intervention component roughly excites attention weights by computing a weighted sum of the attention scores and their absolute values across individual heads.
We apply this approach to the same layer ranges in ours on LLaVA-1.5-7B, Shikra, and MiniGPT-4.
\cref{tab:ablation_head_guide} suggests that our attention intervention method surpasses that of PAI, as our method integrates attention information from various heads to shift attention toward a more truthful direction.

\begin{figure}[t]
    \centering
    \includegraphics[width=1\linewidth]{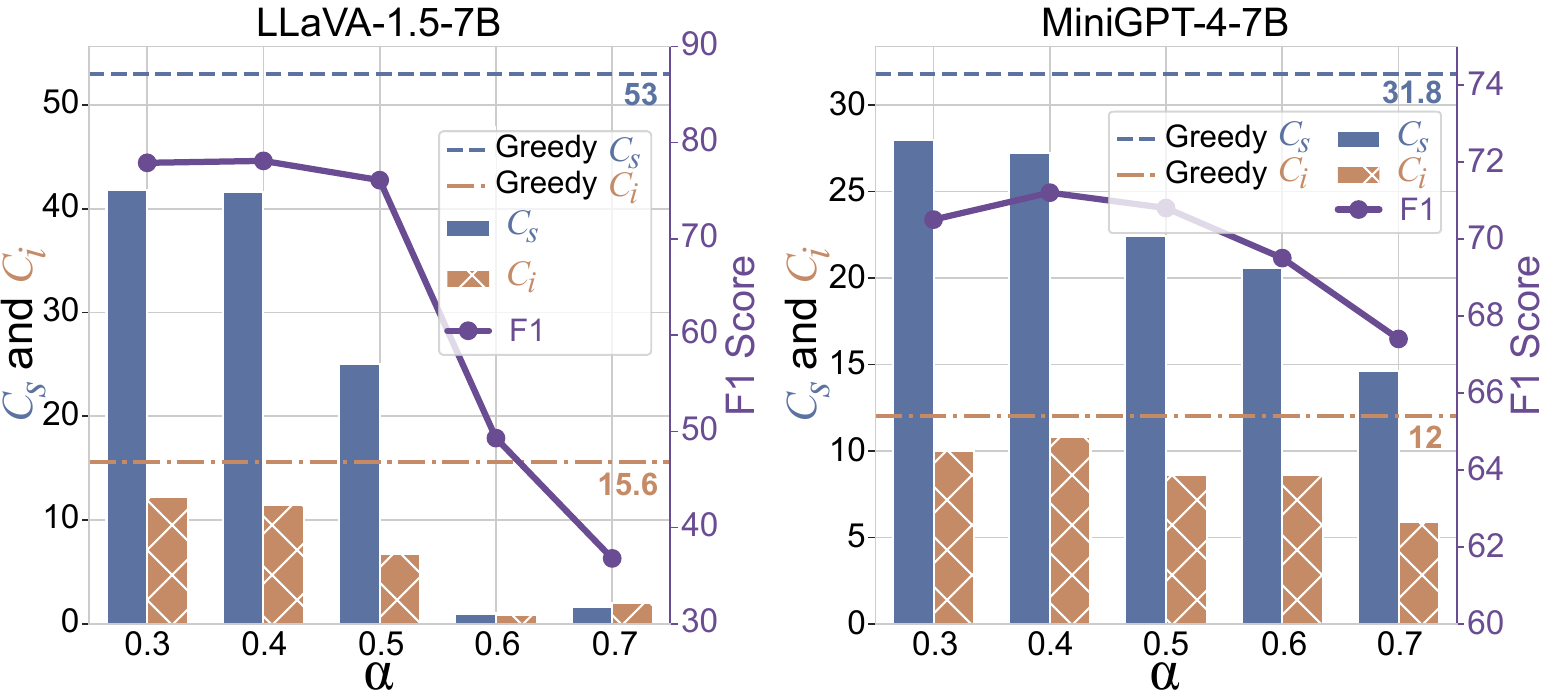}
    \caption{Parameter sensitivity of $\alpha$ with CHAIR metrics.}
    \label{fig:param_ana}
\end{figure}

\begin{table}[t]
    \centering
    \resizebox{.9\linewidth}{!}{
    \large
    \begin{tabular}{lcccccc}
        \toprule
        \multirow{2}{*}{\textbf{Method}} &  \multicolumn{2}{c}{\textbf{LLaVA-1.5-7B}} & \multicolumn{2}{c}{\textbf{Shikra-7B}} & \multicolumn{2}{c}{\textbf{MiniGPT-4-7B}} \\
        \cmidrule{2-3}\cmidrule{4-5}\cmidrule{6-7}
         & $C_S\downarrow$ & $C_I\downarrow$ & $C_S\downarrow$ & $C_I\downarrow$ & $C_S\downarrow$ & $C_I\downarrow$ \\
        \midrule
        PAI$^{\star}$ & 26.6 & 8.1 & 45.8 & 13.6 & 30.2 & 9.5 \\
        \textbf{Ours} & \textbf{25.0} & \textbf{6.7} & \textbf{23.8} & \textbf{9.4} & \textbf{21.4} & \textbf{8.0} \\
        \bottomrule
    \end{tabular}
    }
    \caption{CHAIR hallucination evaluation and F1 results on three LVLMs. $\star$ denotes only using the inference intervention component in the same layer range as ours. Greedy decoding is adopted.}
    \label{tab:ablation_head_guide}
\end{table}

\noindent\textbf{Balance factor $\alpha$ sensitivity.}
To examine the influence of $\alpha$ on intervention efficacy, we vary its value in the range of $\{0.3, 0.4, 0.5, 0.6, 0.7\}$ and the results of CHAIR metrics and F1 score for the 7B version of both LLaVA-1.5 and MiniGPT-4 are shown in~\cref{fig:param_ana}.
We find that a lower $\alpha$ limits hallucination mitigation effectiveness, whereas a higher $\alpha$ compromises the richness of descriptions.
These results indicate that an appropriate $\alpha$ value (\eg, 0.5 for both selected LVLMs) balances the reduction of hallucinations with the maintenance of detailed visual descriptions.

\section{Conclusion and Insights}
We introduce a new perspective, visual information processing, to investigate the underlying mechanisms driving object hallucination. 
Analyzing through the lens of attention, we identify three key findings and discover a simple yet effective method for reducing hallucinations at inference time.
Our study deepens the understanding of how hallucinations are produced in LVLMs.
We conclude with three insights for future direction:
1) Based on LVLMs' interpretation in semantic refinement stage, the image tokens may be further categorized into sub-groups, such as object tokens, color tokens, and textual tokens, enabling us to explore how the LVLMs extract different types of visual information.
2) While we focus on visual information processing, there may exist more granular, stage-wise mechanisms that could explain model behavior more precisely.
3) Instead of external interventions on LVLM outputs, such as contrastive decoding, we suggest that targeted modifications to specific internal states, like attention weights in middle layers, offer a more effective method for calibration.
Our insights may aid the community in developing more trustworthy and reliable LVLMs against hallucinations.

{
\section*{Acknowledgment}
This work is supported by the National Science Foundation of China (62376281, 62206048), the Natural Science Foundation of Jiangsu Province (BK20220819), and the Fundamental Research Funds for the Central Universities (2242024k30035).
}
{
    \small
    \bibliographystyle{ieeenat_fullname}
    \bibliography{main}
}

\appendix
\clearpage
\setcounter{page}{1}
\maketitlesupplementary

\renewcommand{\thesection}{\Alph{section}}
\renewcommand{\thesubsection}{\thesection.\arabic{subsection}}

\section{Limitations}
Despite the simplicity and effectiveness of our hallucination detection and mitigation, there are several limitations:
\begin{itemize}
    \item First, the SVAR metric, used to detect hallucinated object tokens, is limited by the inherent attention behavior of LVLMs. When LVLM consistently exhibits extremely high visual attention ratios at nearly all layers, such as the case of Shikra illustrated in~\cref{fig:shikra_logit_contribution} (a), this may weaken the effectiveness of the SVAR metric.

    \item Second, although the use of the VAR score and logit lens approach can intuitively distinguish two stages of visual information processing, identifying the specific range of these stages remains somewhat subjective. However, leveraging learnable strategies, such as training a set of learnable weights for layers based on the signals from VAR distribution and prediction contributions, could potentially achieve automatic localization of these stages, and we leave this for future work.
\end{itemize}

\begin{table}[b]
    \centering
    \begin{tabular}{lcc}
        \toprule
        \textbf{Model} & \textbf{No. of Real} & \textbf{No. of Hallucinated} \\
        \midrule
        LLaVA-1.5-7B & 4,397 & 1,842 \\
        LLaVA-1.5-13B & 4,488 & 1,700 \\
        Shikra-7B & 4,263 & 1,794 \\
        MiniGPT-4-7B & 2,999 & 981 \\
        \bottomrule
    \end{tabular}
    \caption{Statistical information of case datasets.}
    \label{tab:case_datasets}
\end{table}

\section{Experiment Details}
\subsection{Datasets for Case Study}
\begin{figure}[t]
    \centering
    \includegraphics[width=1\linewidth]{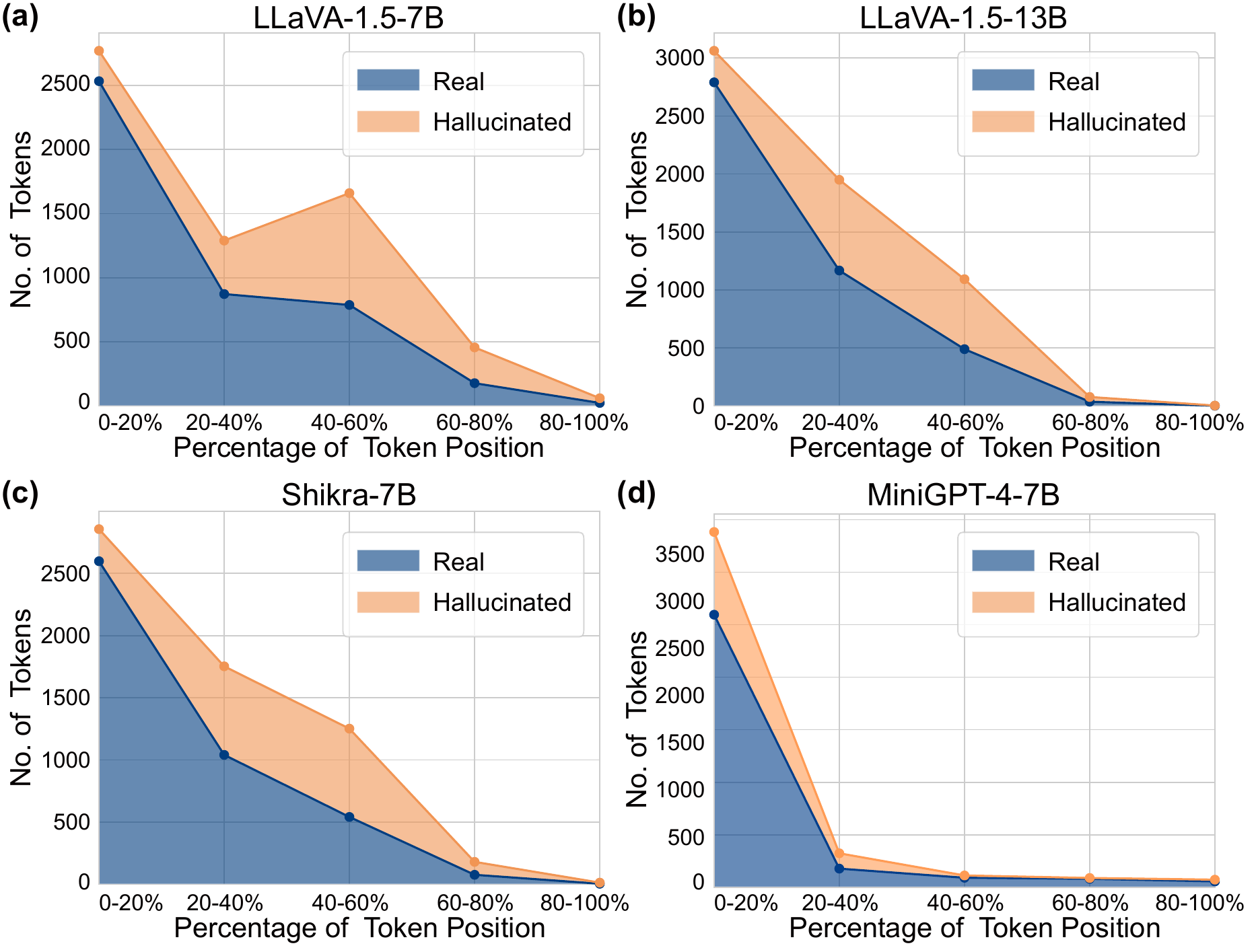}
    \caption{Real and hallucinated object token distributions by their position in description (\%).}
    \label{fig:case_dataset_distribution}
\end{figure}

\cref{tab:case_datasets} reports the statistical information of the synthetic datasets used in our case studies. Additionally, \cref{fig:case_dataset_distribution} illustrates the positional distributions of real and hallucinated object tokens for the four selected LVLMs.

\begin{figure}[t]
    \centering
    \includegraphics[width=1\linewidth]{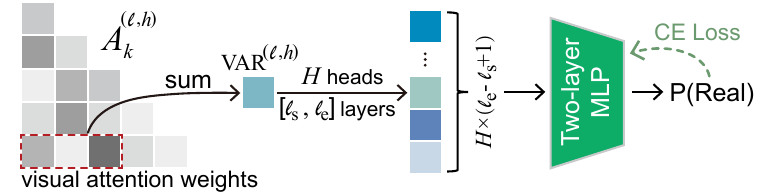}
    \caption{Illustration of detecting hallucinated object tokens by training an MLP classifier on the concatenated VAR scores.}
    \label{fig:detector}
\end{figure}
\begin{table}[t]
    \centering
    \begin{tabular}{lc}
        \toprule
        \textbf{Hyperparameters} & \textbf{LLaVA-1.5-7B} \\
         \midrule
        Optimizer & Adam~\cite{kingma2014adam} \\
        ($\beta_1$, $\beta_2$) & (0.9, 0.999) \\
        Hidden size & \{64, 128, 256, 512\} \\
        Learning rate & \{1e-2, 1e-3, 1e-4\} \\
        No. of epochs & 200 \\
         \bottomrule
    \end{tabular}
    \caption{Training hyperparameters of the two-layer MLP for hallucination detection on LLaVA-1.5-7B.}
    \label{tab:mlp_parameters}
\end{table}

\subsection{MLP Training Details}\label{appendix:detector_train}
\cref{fig:detector} shows the training pipeline of the object hallucination detector.
\cref{tab:mlp_parameters} details the hyperparameters used to train the two-layer MLP, designed for detecting hallucinated object tokens as described in~\cref{subsec:detection}. The Adam optimizer is employed to train the classifier with the number of epochs set to 200. For each layer range, we utilize a gride search strategy to find the optimal hidden layer size and learning rate within the ranges of \{64, 128, 256, 512\} and \{1e-2, 1e-3, 1e-4\}, respectively.

\section{Additional Results}
\subsection{Case Study Results}\label{appendix:case_studies}
In this subsection, we conduct additional experiments on LLaVA-1.5-13B, Shikra-7B, and MiniGPT-4-7B to examine whether other models also share similar characteristics with LLaVA-1.5-7B.

\begin{figure}[t]
    \centering
    \includegraphics[width=1\linewidth]{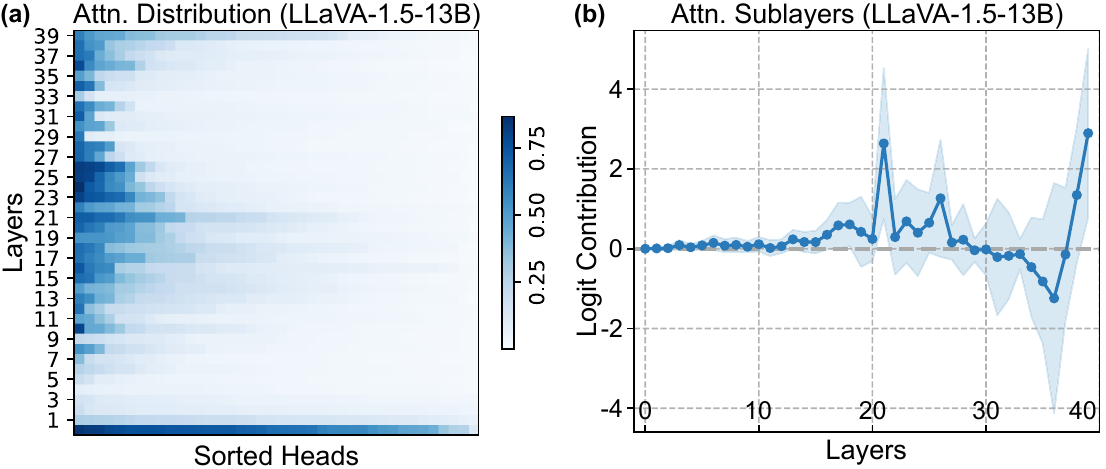}
    \caption{(a) Distribution of visual attention ratio for real object tokens across heads and layers in LLaVA-1.5-13B, sorted row-wise by attention ratios. (b) The logit contribution of attention sublayers to real token prediction.}
    \label{fig:llava13B_var}
\end{figure}

\noindent\textbf{LLaVA-1.5-13B.}~\cref{fig:llava13B_var}~(a)~and~(b) show the VAR score distribution and the prediction contributions from the MHSA sublayers, respectively. We find the same two patterns in the middle layers analogous to those found in LLaVA-1.5-7B as described in~\cref{finding1}, suggesting the model scale generalization of our findings. \cref{fig:llava13b_roc_pr_curve} presents qualitative comparisons of hallucination detection between the $\text{SVAR}_{5\text{-}18}$ metric and the internal confidence method, demonstrating the superiority of our metric.

\begin{figure}[t]
    \centering
    \includegraphics[width=1\linewidth]{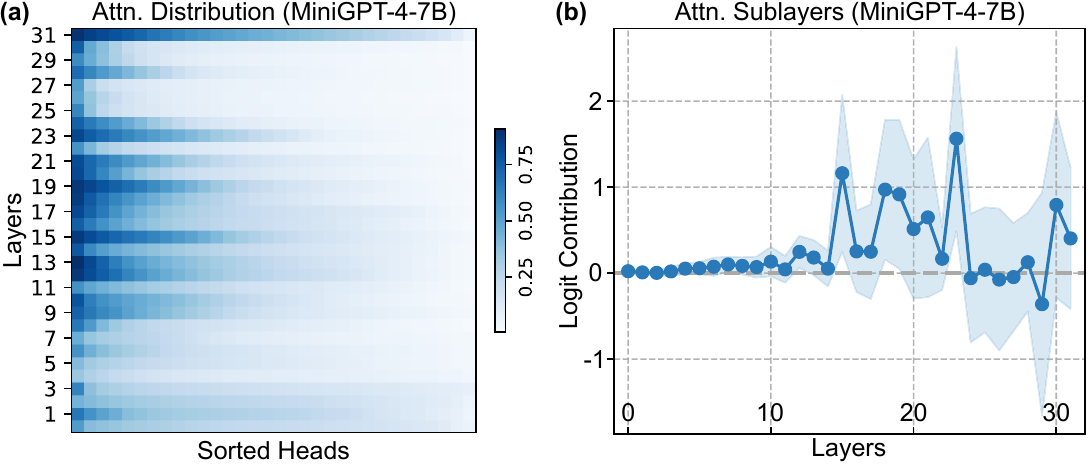}
    \caption{(a) Distribution of visual attention ratio for real object tokens across heads and layers in MiniGPT-4-7B, sorted row-wise by attention ratios. (b) The logit contribution of attention sublayers to real token prediction.}
    \label{fig:minigpt4_logit_contribution}
\end{figure}

\noindent\textbf{MiniGPT-4-7B.}~\cref{fig:minigpt4_logit_contribution}~(a)~and~(b) depict the VAR score distribution and the prediction contributions from the MHSA sublayers, respectively. Similar to LLaVA-1.5-7B, the two patterns in the middle layers where the model exhibits continuous higher visual attention can be observed. Notably, we can see that MiniGPT-4-7B does not exhibit the same high attention as LLaVA-1.5 at the 0-th layer. In our experiments, layers 3-14 are selected as the range of the visual information enrichment stage. \cref{fig:SVAR_distribution_shikra_minigpt}~(b) reports the $\text{SVAR}_{3\text{-}14}$ value distribution across the two token types, demonstrating a similar trend to LLaVA-1.5-7B as described in~\cref{finding2}. These results suggest the model generalization of our findings. The qualitative results of hallucination detection are presented in~\cref{fig:minigpt_curves}.

\begin{figure}[t]
    \centering
    \includegraphics[width=1\linewidth]{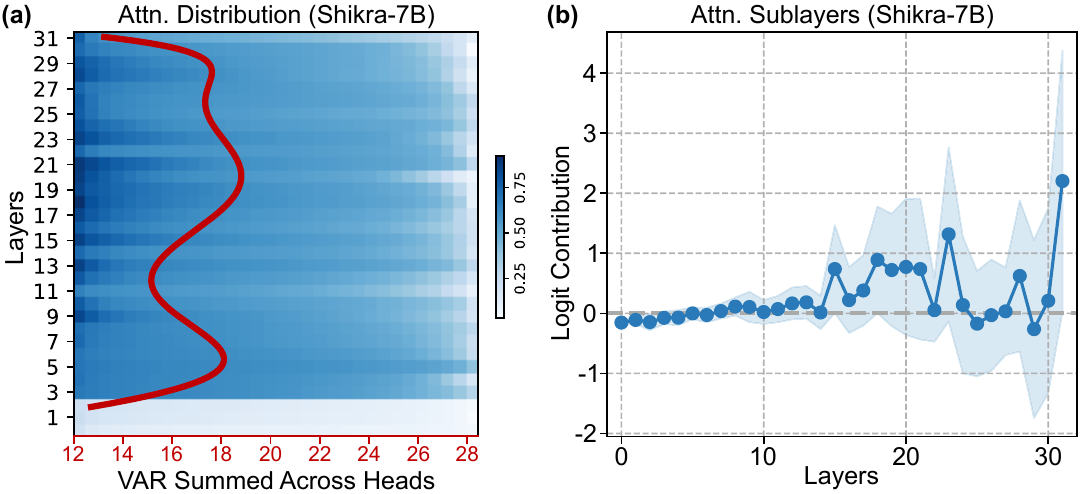}
    \caption{(a) Distribution of visual attention ratio for real object tokens across heads and layers in Shikra-7B, sorted row-wise by attention ratios. Note that the red curve represents a seventh-order polynomial fit to the values of attention ratios summed over heads in each layer. (b) The logit contribution of attention sublayers to real token prediction.}
    \label{fig:shikra_logit_contribution}
\end{figure}

\begin{figure}[t]
    \centering
    \includegraphics[width=1\linewidth]{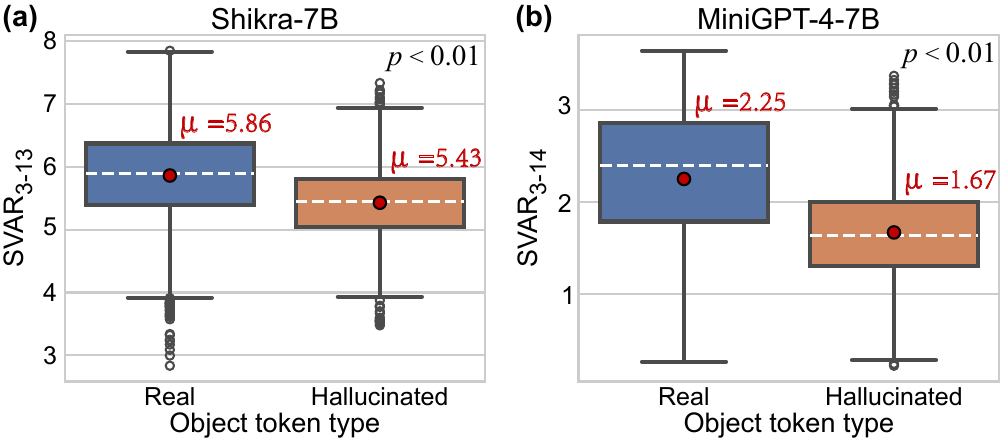}
    \caption{$\text{SVAR}_{3\text{-}13}$ and $\text{SVAR}_{3\text{-}14}$ score distributions across object token types for Shikra-7B (a) and MiniGPT-4-7B (b), respectively.}
    \label{fig:SVAR_distribution_shikra_minigpt}
\end{figure}

\begin{figure}[t]
    \centering
    \includegraphics[width=1\linewidth]{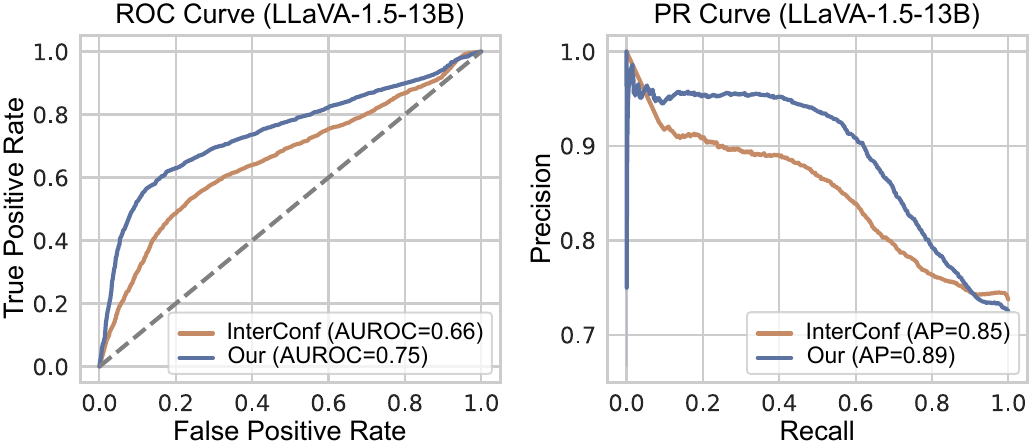}
    \caption{Object hallucinations detection curves for LLaVA-1.5-13B.}
    \label{fig:llava13b_roc_pr_curve}
\end{figure}

\begin{figure}[t]
    \centering
    \includegraphics[width=1\linewidth]{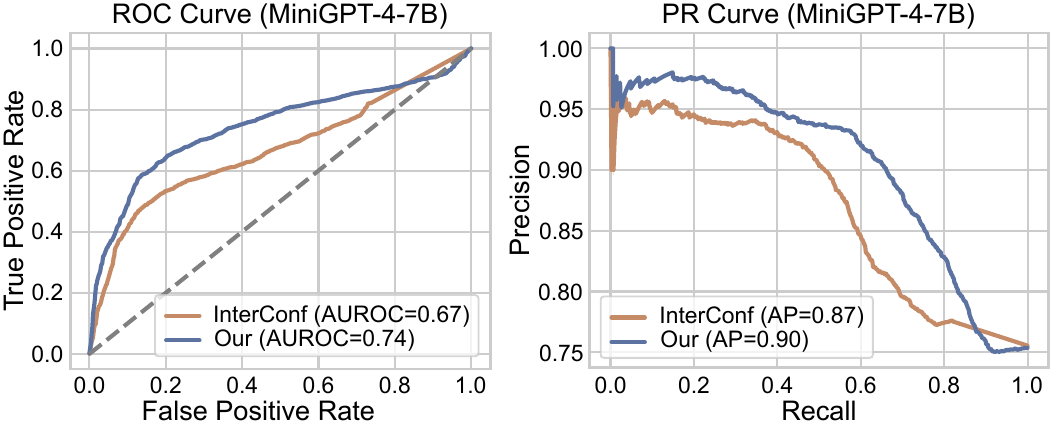}
    \caption{Object hallucinations detection curves for MiniGPT-4-7B.}
    \label{fig:minigpt_curves}
\end{figure}

\begin{figure}[t]
    \centering
    \includegraphics[width=1\linewidth]{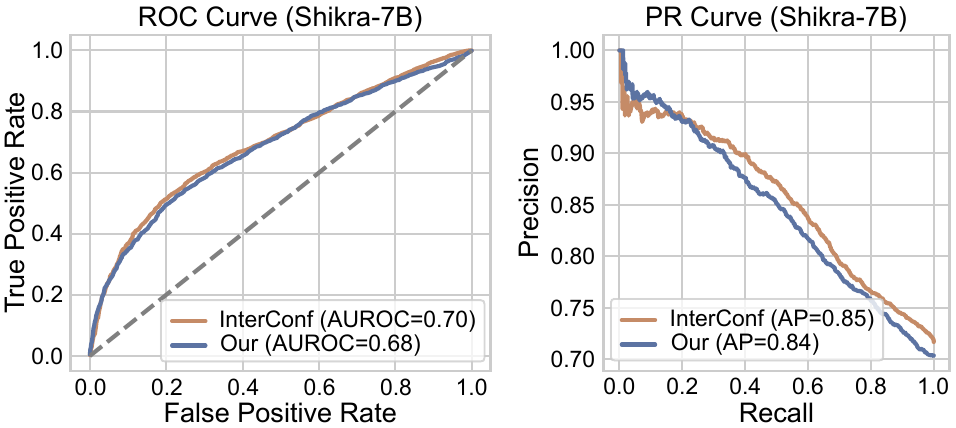}
    \caption{Object hallucinations detection curves for Shikra-7B.}
    \label{fig:shikra_curves}
\end{figure}

\noindent\textbf{Shikra-7B.} As shown in~\cref{fig:shikra_logit_contribution}~(a), Shikra continuously exhibits extremely high VAR scores across layers. To clearly analyze the VAR distribution, a seventh-order polynomial is used to fit the summed VAR values over all heads of each layer (depicted by a red curve). Compared to other layers, we can see that the middle layers exhibit relatively higher VAR scores, aligning with our observation from LLaVA-1.5-7B. Combined with the prediction contributions from MHSA sublayers in~\cref{fig:shikra_logit_contribution}~(b), we can also identify two distinct patterns in the middle layers. Like MiniGPT-4-7B, Shikra-7B exhibits low visual attention at the 0-th layer. In our experiments, layers 3-13 are selected as the range of the visual information enrichment stage. \cref{fig:SVAR_distribution_shikra_minigpt}~(a) presents the $\text{SVAR}_{3\text{-}13}$ value distribution across the two token types, demonstrating a similar trend to other LVLMs. The comparison results of hallucination detection, displayed in~\cref{fig:shikra_curves}, show that our simple $\text{SVAR}_{3\text{-}13}$ metric performs comparably to the more complex baseline that projects the hidden states of all image tokens at all layers into the vocabulary space.
Compared to other LVLMs, the decreased performance of the SVAR metric on Shikra-7B may be attributed to the extremely high VAR scores across nearly all layers, potentially reducing the sensitivity of our metric to attention pattern differences.

\subsection{Results of Statistical Tests}\label{appendix:statistical_test}
To assess the statistical significance of the SVAR score being higher for real object tokens than for hallucinated ones during visual information enrichment, we conduct a one-tailed t-test for each LVLM. We present the results in~\cref{tab:statistical_test_llava7B} for LLaVA-1.5-7B,~\cref{tab:statistical_test_llava13B} for LLaVA-1.5-13B,~\cref{tab:statistical_test_shikra} for Shikra-7B, and~\cref{tab:statistical_test_minigpt} for MiniGPT-4-7B.
Across all models, the results consistently indicate that significantly higher attention weights are assigned to image tokens when generating real object tokens, compared to hallucinated ones.


\subsection{Numerical Results of $\alpha$ Sensitivity}
\cref{tab:ablation_param_sensitivity} presents the sensitivity results of the balance factor $\alpha$, used in our attention intervention method (\cref{eq:enhance}), on LLaVA-1.5-7B, LLaVA-1.5-13B, and MiniGPT-4-7B. In addition to modulating the trade-off between hallucination mitigation and description richness as discussed in~\cref{ablation}, we find that LLaVA-1.5-7B and LLaVA-1.5-13B are more sensitive to changes in $\alpha$ compared to MiniGPT-4-7B. A possible reason for this increased sensitivity may be that LLaVA-1.5 uses substantially more image tokens than MiniGPT-4 (576 versus 32), potentially magnifying the impact of the parameter $\alpha$.

\begin{table}[t]
    \centering
    \begin{tabular}{rcc}
        \toprule
        \textbf{LLaVA-1.5-7B} & \textbf{Real} & \textbf{Hallucinated}\\
        \midrule
        $\textrm{SVAR}_{5\textrm{-}18}$ score & 1.70 & 1.25 \\
        t-statistic & \multicolumn{2}{c}{32.44} \\
        p-value & \multicolumn{2}{c}{1.03E-213} \\
        df & \multicolumn{2}{c}{6,237} \\
        \bottomrule
    \end{tabular}
    \caption{Results of one-tailed t-tests for LLaVA-1.5-7B. The null hypothesis is the mean $\textrm{SVAR}_{5\textrm{-}18}$ score of real object tokens is less than or equal to the mean $\textrm{SVAR}_{5\text{-}18}$ score of hallucinated ones.}
    \label{tab:statistical_test_llava7B}
\end{table}
\begin{table}[t]
    \centering
    \begin{tabular}{rcc}
        \toprule
        \textbf{LLaVA-1.5-13B} & \textbf{Real} & \textbf{Hallucinated}\\
        \midrule
        $\textrm{SVAR}_{5\textrm{-}18}$ score & 1.50 & 1.06 \\
        t-statistic & \multicolumn{2}{c}{32.24} \\
        p-value & \multicolumn{2}{c}{3.25E-211} \\
        df & \multicolumn{2}{c}{6,186} \\
        \bottomrule
    \end{tabular}
    \caption{Results of one-tailed t-tests for LLaVA-1.5-13B. The null hypothesis is the mean $\textrm{SVAR}_{5\text{-}18}$ score of real object tokens is less than or equal to the mean $\textrm{SVAR}_{5\textrm{-}18}$ score of hallucinated ones.}
    \label{tab:statistical_test_llava13B}
\end{table}
\begin{table}[t]
    \centering
    \begin{tabular}{rcc}
        \toprule
        \textbf{Shikra-7B} & \textbf{Real} & \textbf{Hallucinated}\\
        \midrule
        $\textrm{SVAR}_{3\text{-}13}$ score & 5.86 & 5.43 \\
        t-statistic & \multicolumn{2}{c}{22.56} \\
        p-value & \multicolumn{2}{c}{1.50E-108} \\
        df & \multicolumn{2}{c}{6,055} \\
        \bottomrule
    \end{tabular}
    \caption{Results of one-tailed t-tests for Shikra-7B. The null hypothesis is the mean $\textrm{SVAR}_{3\text{-}13}$ score of real object tokens is less than or equal to the mean $\textrm{SVAR}_{3\textrm{-}13}$ score of hallucinated ones.}
    \label{tab:statistical_test_shikra}
\end{table}
\begin{table}[t]
    \centering
    \begin{tabular}{rcc}
        \toprule
        \textbf{MiniGPT-4-7B} & \textbf{Real} & \textbf{Hallucinated}\\
        \midrule
        $\textrm{SVAR}_{3\text{-}14}$ score & 2.25 & 1.67 \\
        t-statistic & \multicolumn{2}{c}{22.06} \\
        p-value & \multicolumn{2}{c}{4.21E-102} \\
        df & \multicolumn{2}{c}{3,978} \\
        \bottomrule
    \end{tabular}
    \caption{Results of one-tailed t-tests for MiniGPT-4-7B. The null hypothesis is the mean $\textrm{SVAR}_{3\text{-}14}$ score of real object tokens is less than or equal to the mean $\textrm{SVAR}_{3\textrm{-}14}$ score of hallucinated ones.}
    \label{tab:statistical_test_minigpt}
\end{table}

\begin{table}[t]
    \centering
    \resizebox{\linewidth}{!}{
    \large
    \begin{tabular}{lccccccccc}
        \toprule
        \multirow{2}{*}{\textbf{\Large$\mathbf{\alpha}$}} &  \multicolumn{3}{c}{\textbf{LLaVA-1.5-7B}} & \multicolumn{3}{c}{\textbf{LLaVA-1.5-13B}} & \multicolumn{3}{c}{\textbf{MiniGPT-4-7B}}  \\
        \cmidrule{2-4}\cmidrule{5-7}\cmidrule{8-10}
         & $C_S\downarrow$ & $C_I\downarrow$ & F1$\uparrow$ & $C_S\downarrow$ & $C_I\downarrow$ & F1$\uparrow$ & $C_S\downarrow$ & $C_I\downarrow$ & F1$\uparrow$ \\
        \midrule
        Greedy & 53.0 & 15.6 & \cellcolor{gray!15}76.7 & 49.8 & 14.6 & \cellcolor{gray!15}78.2 & 31.8 & 12.0 & \cellcolor{gray!15}71.1 \\
        \midrule
        0.3 & 41.8 & 12.2 & \cellcolor{gray!15}77.9 & 44.4 & 12.5 & \cellcolor{gray!15}78.1 & 28.0 & 10.0 & \cellcolor{gray!15}70.5 \\
        0.4 & 41.6 & 11.5 & \cellcolor{gray!15}78.1 & 44.6 & 13.2 & \cellcolor{gray!15}77.5 & 27.2 & 10.8 & \cellcolor{gray!15}71.2 \\
        0.5 & 25.0 & 6.7 & \cellcolor{gray!15}76.1 & 25.8 & 8.8 & \cellcolor{gray!15}77.3 & 22.4 & 8.6 & \cellcolor{gray!15}70.8 \\
        0.6 & 1.0 & 0.9 & \cellcolor{gray!15}49.3 & 6.4 & 3.3 & \cellcolor{gray!15}57.7 & 20.6 & 8.6 & \cellcolor{gray!15}69.5 \\
        0.7 & 1.6 & 2.0 & \cellcolor{gray!15}36.8 & 2.4 & 26.3 & \cellcolor{gray!15}40.0 & 14.6 & 5.9 & \cellcolor{gray!15}67.4 \\
        \bottomrule
    \end{tabular}
    }
    \caption{Numerical results of balance factor $\alpha$ sensitivity.}
    \label{tab:ablation_param_sensitivity}
\end{table}


\begin{table}[t]
    \centering
    \resizebox{1\linewidth}{!}{
    \begin{tabular}{lcccccc}
    \toprule
    \textbf{LLaVA-1.5-7B} & Greedy & Beam & OPERA & VCD$^{\dagger}$ & PAI$^{\dagger}$ & \textbf{Ours} \\
    \midrule
    \textbf{CHAIR$\downarrow$} & 7.7 & 9.1 & 7.3 & 8.6 & 4.9  & \textbf{4.3} \\
    \textbf{Hal$\downarrow$} & 35.4  & 39.8 & 31.5 & 39.5 & 24.4 & \textbf{20.2} \\
    \textbf{Cog$\downarrow$} & 4.3  & 4.8 & 2.9 & 4.5 & 1.6 & \textbf{1.2} \\
    \bottomrule
    \end{tabular}
    }
    \caption{AMBER results on LLaVA-1.5-7B with \textit{max new token} set to 512. $\dagger$ denotes using the greedy decoding strategy.}\label{tab:amber_res}
\end{table}

\subsection{Attention Heads Behavior Visualization}\label{appendix:head_behavior}
\begin{figure*}
    \centering
    \includegraphics[width=.87\linewidth]{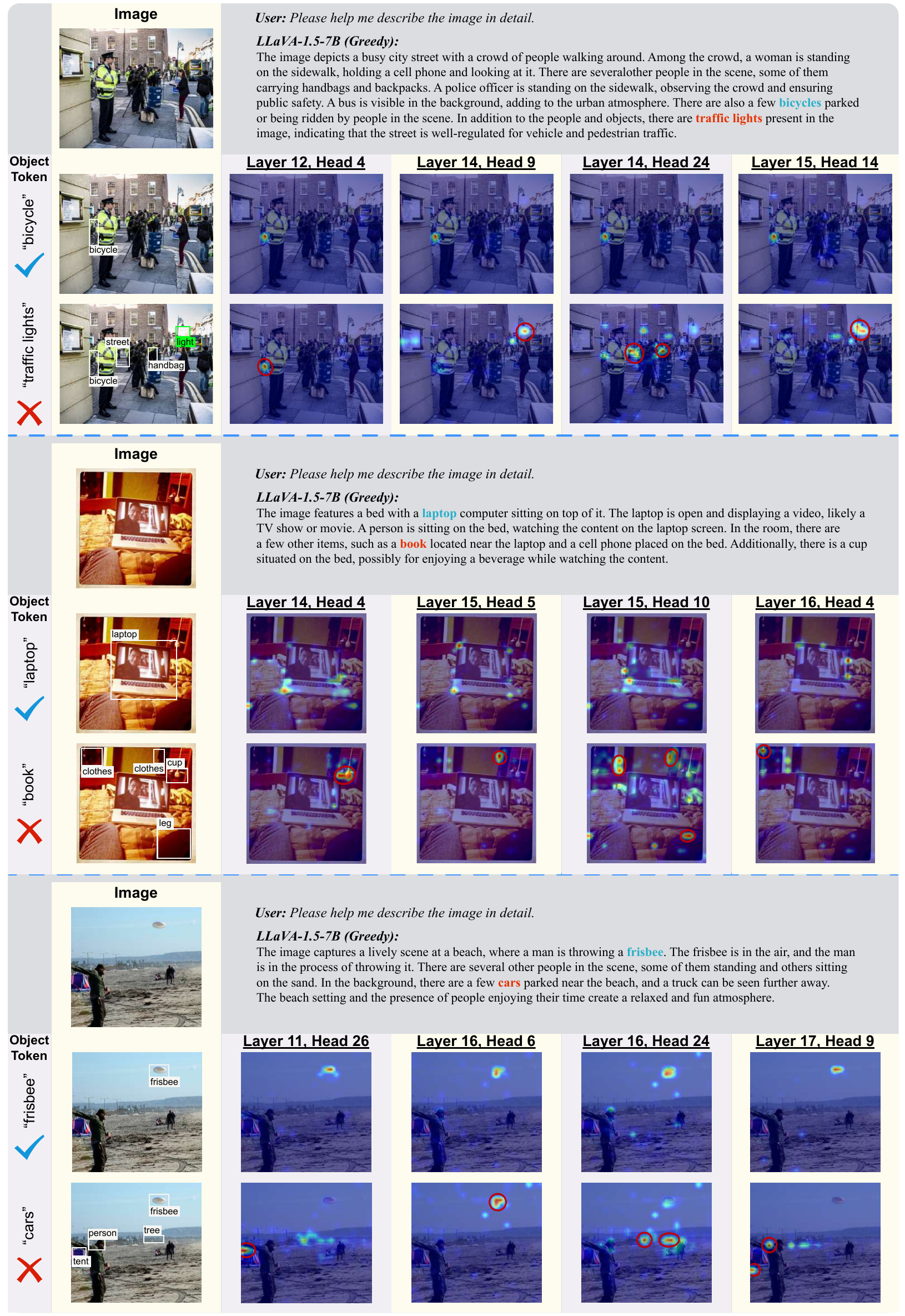}
    \caption{Attention heads behavior in LLaVA-1.5-7B.}
    \label{app_fig:heads_heatmap1}
\end{figure*}

\begin{figure*}
    \centering
    \includegraphics[width=.87\linewidth]{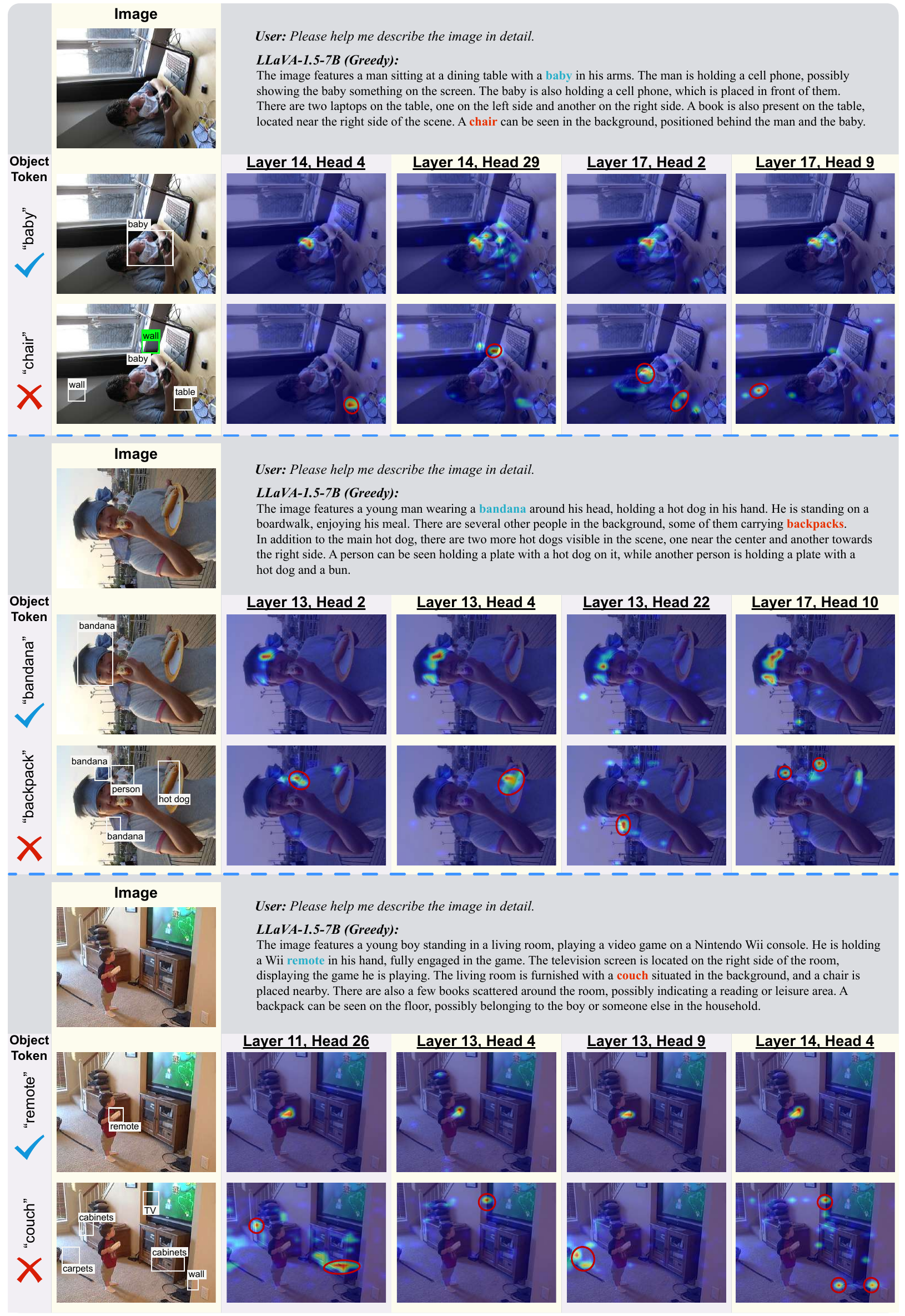}
    \caption{Attention heads behavior in LLaVA-1.5-7B.}
    \label{app_fig:heads_heatmap2}
\end{figure*}

We exhibit more visualization examples of LLaVA-1.5-7B in~\cref{app_fig:heads_heatmap1,app_fig:heads_heatmap2} to validate that the heads interact with inconsistent objects in the image during visual information enrichment when generating hallucinated object tokens.

\subsection{Comparison Results on AMBER Benchmark}
We further evaluate our approach on AMBER~\cite{wang2023llm} benchmark, which contains 1,004 images for the generative task.
The results presented in~\cref{tab:amber_res} demonstrate its superior performance.

\subsection{Some Intuition Behind Layer Division}
The layers of LLaVA-1.5-7B are divided into four ranges according to the patterns identified from the results of VAR score (\cref{fig1:attention}~(a)) and logit lens (\cref{fig1:attention}~(b) and \cref{fig2:logit_lens}).
We find \textbf{Range 1} (layer 0-4): low-level image processing, VAR attention pattern differs in different models;
\textbf{Range 2} (layer 5-18): visual information enrichment, accumulates the visual information exhibiting high VAR scores and low logit contribution;
\textbf{Range 3} (layer 19-26): semantic refinement, interacts semantic information of image tokens with high VAR scores and reasons object token prediction with high logit contribution;
\textbf{Range 4} (layer 27-31): grammar concern, guarantees the coherence and correctness of response with low VAR scores, in which the model tends to interpret image tokens as punctuation marks or conjunctions (\cref{fig2:logit_lens}).
These patterns can be generalized to divide other LVLMs similarly.
It is worth noting that the first and last divisions are not fixed at 5 layers but vary with models.

\subsection{Qualitative Results of Hallucination Mitigation}\label{appendix:qualitative_results}
\begin{figure*}
    \centering
    \includegraphics[width=1\linewidth]{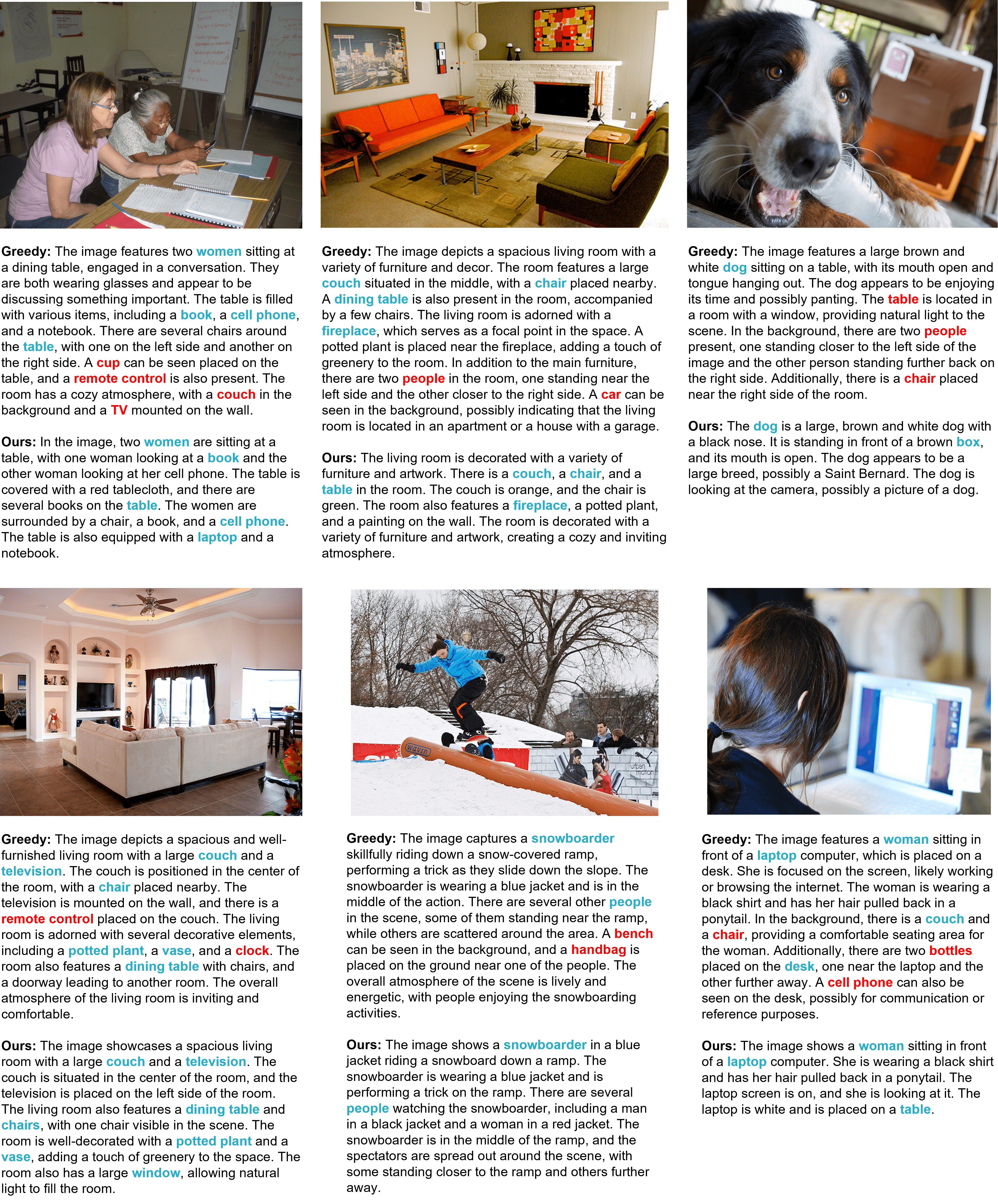}
    \caption{Qualitative results of hallucination mitigation on LLaVA-1.5-7B. The real and hallucinated object words are marked in blue and red, respectively. Our prompt is ``Please help me describe the image in detail.''.}
    \label{app_fig:llava_case}
\end{figure*}

\begin{figure*}
    \centering
    \includegraphics[width=1\linewidth]{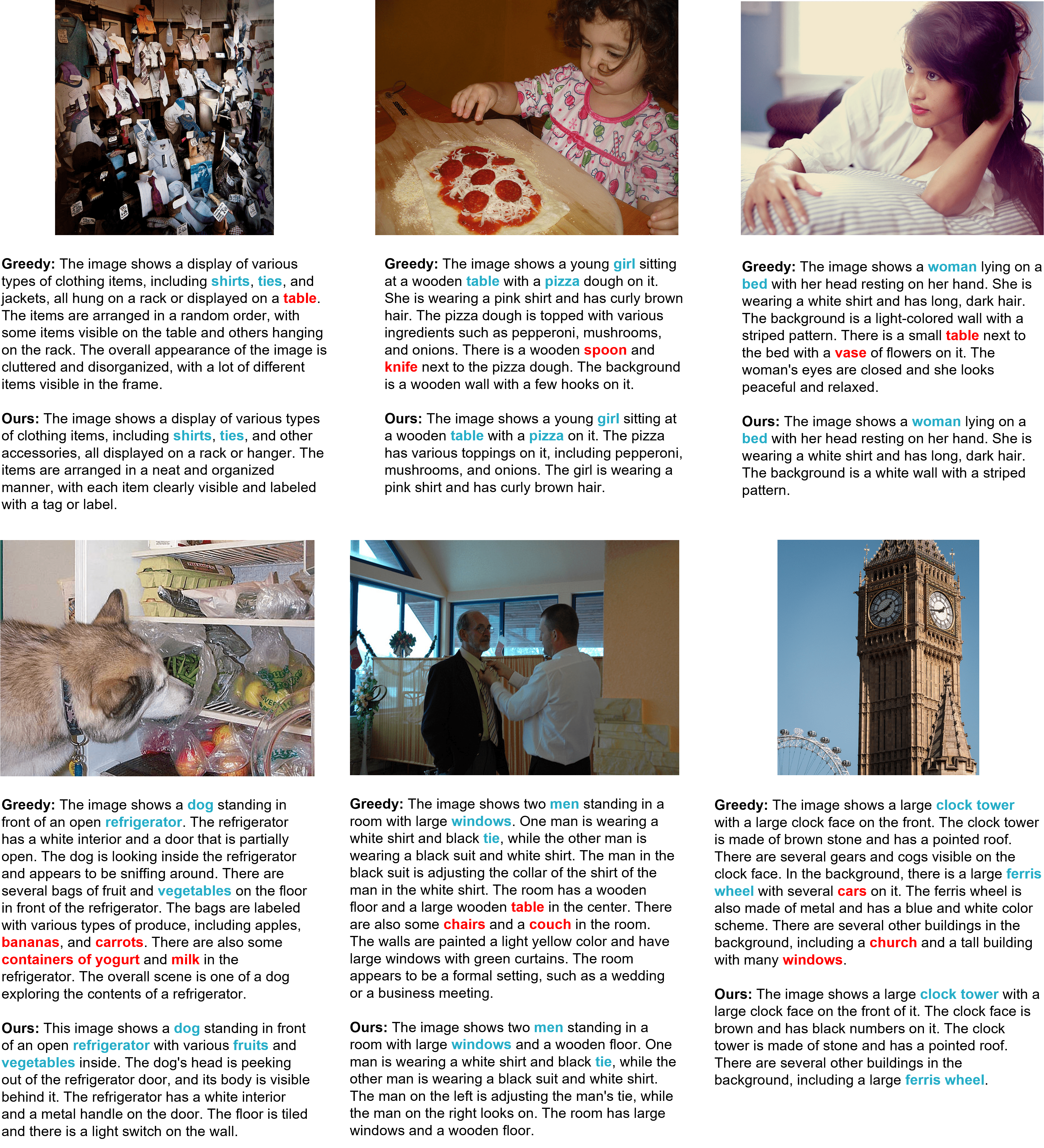}
    \caption{Qualitative results of hallucination mitigation on MiniGPT-4-7B. The real and hallucinated object words are marked in blue and red, respectively. Our prompt is ``Please help me describe the image in detail.''.}
    \label{app_fig:minigpt_case}
\end{figure*}

\begin{figure*}
    \centering
    \includegraphics[width=1\linewidth]{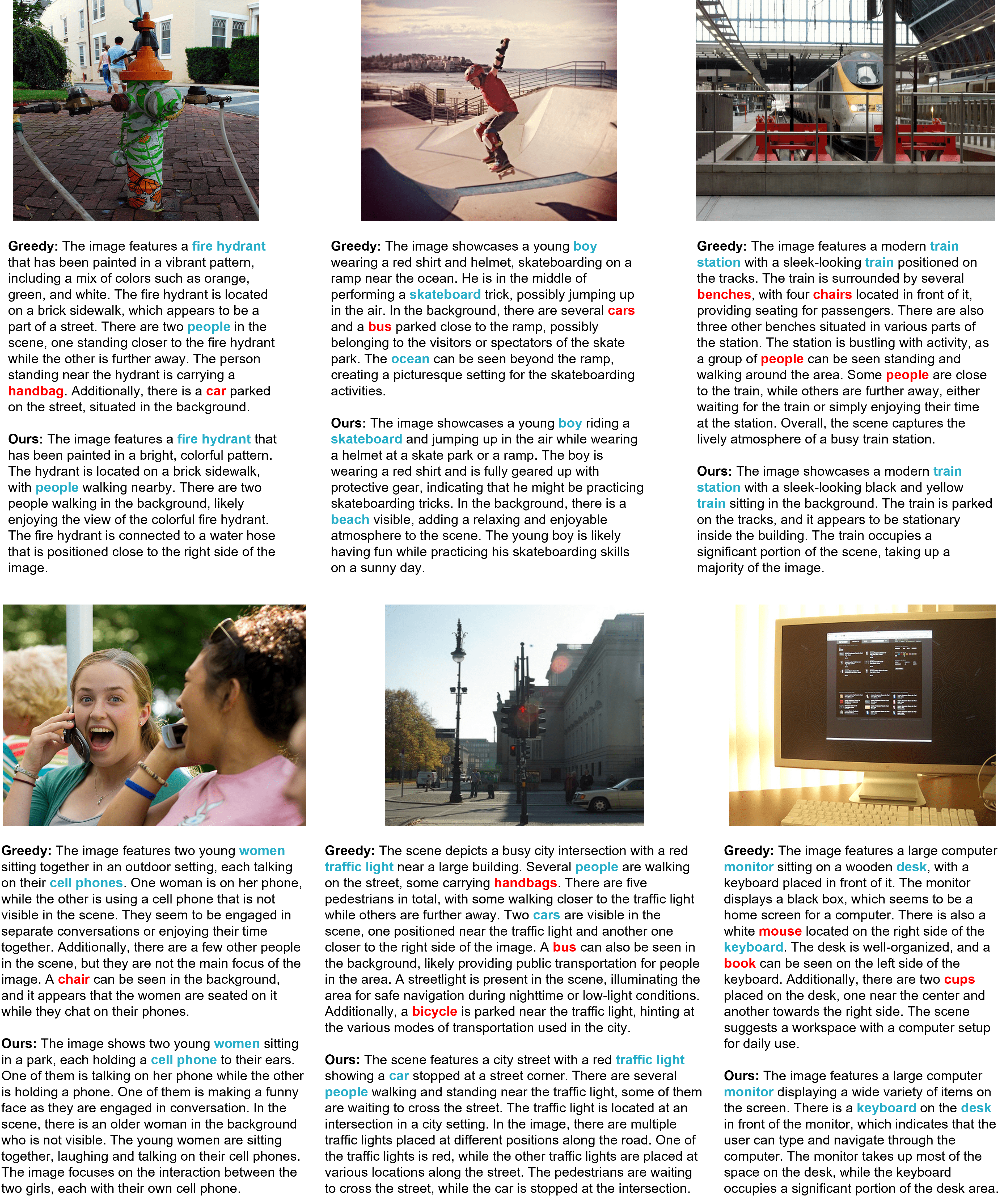}
    \caption{Qualitative results of hallucination mitigation on Shikra-7B. The real and hallucinated object words are marked in blue and red, respectively. Our prompt is ``Please help me describe the image in detail.''.}
    \label{app_fig:shikra_case}
\end{figure*}

We provide sample results from our hallucination mitigation method as described in~\cref{subsec:method}, which corrects the attention distribution over image tokens, in~\cref{app_fig:llava_case},~\cref{app_fig:minigpt_case} and~\cref{app_fig:shikra_case} for LLaVA-1.5-7B, MiniGPT-4-7B, and Shikra-7B, respectively.

\newpage

\end{document}